%% file: cayopt.tex
\documentclass[shortAfour,sageh,times]{sagej}
\usepackage{moreverb,url}
\usepackage[colorlinks,bookmarksopen,bookmarksnumbered,citecolor=red,urlcolor=red]{hyperref}

\usepackage[printonlyused,withpage]{acronym}
\usepackage{amscd}

\hypersetup{
    pdftitle={},
    pdfauthor={},
    pdfkeywords={},
    pdfsubject={},
    pdfstartview=FitH,
    bookmarks=true,
    bookmarksopen=true,
    breaklinks=true,
    colorlinks=true,
    linkcolor=black,
    anchorcolor=black,
    citecolor=black,
    filecolor=black,
    menucolor=black,
    pagecolor=black,
    urlcolor=black
}

\graphicspath{{figs/}}
\input{notation_header}

\acrodef{BA}{Bundle Adjustment}
\acrodef{SLAM}{Simultaneous Localization and Mapping}
\acrodef{RANSAC}{Random Sample And Consensus}
\acrodef{QCQP}{Quadratically Constrained Quadratic Program}
\acrodef{SDP}{Semidefinite Program}
\acrodef{IRLS}{Iteratively Reweighted Least-Squares}
\acrodef{TLS}{Total Least-Squares}
\acrodef{SVD}{Singular-Value Decomposition}
\acrodef{GM}{Geman-McClure}
\acrodef{PSD}{positive-semidefinite}
\acrodef{GP}{Gaussian Process}
\acrodef{WNOA}{White Noise on Acceleration}
\acrodef{LICQ}{Linearly Independent Constraint Qualification}
\acrodef{SVR}{Singular Value Ratio}

\hypersetup{%
    pdftitle={},
    pdfauthor={},
    pdfkeywords={},
    pdfsubject={},
    pdfstartview=FitH,%
    bookmarks=true,%
    bookmarksopen=true,%
    breaklinks=true,%
    colorlinks=true,%
    linkcolor=black,
    anchorcolor=black,%
    citecolor=black,
    filecolor=black,%
    menucolor=black,
    pagecolor=black,%
    urlcolor=black
}%

\newcommand{\change}[1]{\color{black}#1\color{black}}

\begin{document}

\runninghead{Barfoot et al.}

\title{Certifiably Optimal Rotation and Pose Estimation Based on the Cayley Map}

\author{Timothy D. Barfoot\affilnum{1}, Connor Holmes\affilnum{1}, and Frederike D\"{u}mbgen\affilnum{1}}
\affiliation{\affilnum{1}Robotics Institute, University of Toronto}
\corrauth{Timothy D. Barfoot,
Institute for Aerospace Studies,
University of Toronto,
Toronto, Ontario,
M3H~5T6, Canada.}
\email{tim.barfoot@utoronto.ca}

\begin{abstract}
We present novel, convex relaxations for rotation and pose estimation problems that can \change{{\em a posteriori} guarantee global optimality for practical measurement noise levels. }   Some such relaxations exist in the literature for specific problem setups that assume the matrix von Mises-Fisher distribution (a.k.a., matrix Langevin distribution or chordal distance) for isotropic rotational uncertainty.  However, another common way to represent uncertainty for rotations and poses is to define anisotropic noise in the associated Lie algebra.  Starting from a noise model based on the Cayley map, we define our estimation problems, convert them to \acp{QCQP}, then relax them to \acp{SDP}, which can be solved using standard interior-point optimization methods\change{; global optimality follows from Lagrangian strong duality. }  We first show how to carry out basic rotation and pose averaging.  We then turn to the more complex problem of trajectory estimation, which involves many pose variables with both individual and inter-pose measurements (or motion priors).  Our contribution is to formulate \ac{SDP} relaxations for all these problems \change{based on the Cayley map (including the identification of redundant constraints) and to show them working in practical settings}. We hope our results can add to the catalogue of useful estimation problems whose \change{solutions can be {\em a posteriori} guaranteed to be globally optimal.}
\end{abstract}


\maketitle

\section{Introduction}

{\em State estimation} is concerned with fusing several noisy measurements (and possibly a prior model) into a less noisy estimate of the state (e.g., position, velocity, orientation) of a vehicle, robot, or other object of interest.  Real-world state estimation problems often involve measurement functions and motion models that are nonlinear with respect to the state.  Alternatively, the state itself may not be an element of a vector space, such as the rotation of a rigid body.  These challenging aspects typically mean that when we set up our estimator as an optimization, it is a {\em nonconvex} problem; the cost function, the feasible set, or both are not convex.  Nonconvex optimization problems are in general much harder to solve than convex ones because they can have local minima and common gradient-based optimizers can easily become trapped therein.  

For example, we might have a generic nonlinear least-squares problem such as
\begin{equation}
\underset{\mbf{x}}{\mbox{min}} \sum_{m=1}^M \left( \mbf{y}_m - \mbf{g}(\mbf{x}) \right)^T \left( \mbf{y}_m - \mbf{g}(\mbf{x}) \right),
\end{equation}
where $\mbf{x}$ is the unknown state, $\mbf{y}_m$ are noisy measurements, and $\mbf{g}(\cdot)$ is a measurement function.  If $\mbf{g}(\mbf{x})$ is linear in $\mbf{x}$, then this problem is convex, but otherwise it often is not.  Using gradient descent or Gauss-Newton to solve this problem means we usually require a good initial guess for $\mbf{x}$ to arrive at the global minimum.  What if such an initial guess is not available?  Could we solve a problem such as this one globally without such a guess?  It turns out the answer may be yes, depending on the specific problem to be solved.

There has been quite a bit of work in robotics and computer vision aimed at the idea of solving estimation problems globally.  Most of these works employ sophisticated tools from the optimization literature to achieve this.  In particular, {\em Lagrangian duality} is used to derive {\em convex relaxations}, which can be solved globally.  \citet[\S 5]{boyd04} provide the necessary background on duality theory.  We will be following a common pathway where we first convert our nonconvex optimization problem into another nonconvex form called a \acf{QCQP}; from here we relax this to a (convex) \acf{SDP}, amenable to off-the-shelf solvers (e.g., the interior-point-based \ac{SDP} solver in \texttt{mosek} \citep{mosek}).  \change{This last step is known as {\em Shor's relaxation} \citep{shor87}. } Theoretically, \acp{SDP} admit polynomial-time solutions because they are convex; in practice, modern \ac{SDP} problems with a few thousand variables can be solved in reasonable time but struggle to scale up beyond this.  Our contribution in this paper is to show that we can solve a set of estimation problems involving rotations and poses globally, using these convex relaxation tools; the novelty lies in the fact that our particular problems \change{(formulated using the Cayley map) } have not been examined in this globally optimal framework before.     

As mentioned, the convex relaxation procedure we employ \change{has been well known in the optimization community for some time. In particular, it has been used for {\em polynomial optimization} \citep{parrilo03} and in various {\em combinatorial optimization} problems, such as {\em quadratic assignment} \citep{nesterovSemidefiniteProgrammingRelaxations2000} and {\em max-cut graph partitioning} \citep{anjosStrengthenedSemideyniteRelaxations2002}. }
\change{To the authors' knowledge, the first use of \ac{SDP} relaxations in the robotics community was by  \citet{liuConvexOptimizationBased2012} for {\em planar \ac{SLAM}}, though their application in computer vision \citep{kahlGloballyOptimalEstimates2005} and signal processing \citep{luoSemidefiniteRelaxationQuadratic2010a,krislockExplicitSensorNetwork2010} occurred earlier.  More generally, }\citet{cifuentes22} provides a nice overview of some common problems in computer vision and robotics where these tools have been applied before, as well as providing a rationale for why they are so effective.
\change{One of the most commonly investigated relaxations in the robotics and vision communities is } {\em rotation synchronization}\footnote{Some authors instead use the term `rotation averaging', whereas we use rotation averaging to mean the process of fusing several noisy measurements of a single rotation into an estimate.} \citep{bandeira17, eriksson18, eriksson19, dellaert20}; here several rotations are linked through noisy relative rotation measurements.  Rotation synchronization turns out to be the nucleus of several of the other problems under study including {\em pose-graph optimization} \citep{briales17, carlone18, rosen19, tian21,carlonePlanarPoseGraph2016}, {\em point-set registration} \citep{chaudhury15, iglesias20, yang20}, {\em calibration} \citep{giamou19,wise20}, \change{{\em mutual localization} \citep{wangCertifiablyOptimalMutual2022}, } and {\em landmark-based \ac{SLAM}} \citep{holmes23}.   Rotation synchronization and its cousins have been shown to admit fast solutions through low-rank factorizations \citep{rosen19}.  More recently, other measurement models such as {\em range sensing} have also been incorporated into globally certifiable problems \citep{papalia22,dumbgen22}.  

 \change{It is worth also mentioning the works of \citep{horowitz2014convex, forbes2015linear} on globally optimal pointcloud alignment, which also employ convex relaxation, but the route to the \ac{SDP} is not Shor's relaxation; instead, Lie group optimization variables are relaxed to live in a convex set.  On the surface, this approach is not applicable to the problems considered herein; for example, our cost functions are not always initially convex.}

A common thread that ties most of the existing literature together is that the {\em chordal distance} is used to construct the terms of the cost function that involve rotation variables.  Viewed through a probabilistic lens, the chordal distance is related to the matrix Langevin or matrix von Mises-Fisher distribution, whose density can be written in the form
\begin{equation}
p(\mbf{C}) \propto \exp\left( -\frac{1}{2}\sigma^{-1} \, \mbox{tr}\left((\mbf{C}-\bar{\mbf{C}})(\mbf{C}-\bar{\mbf{C}})^T\right) \right),
\end{equation}
where $\mbf{C}$ is \change{a rotation matrix}, $\bar{\mbf{C}}$ is the mode, and $\sigma > 0$ is a scalar concentration parameter; this distribution is isotropic, which is one limitation we aim to overcome in this work.
This is not the only way to represent rotational uncertainty.  Another common way is to use exponential coordinates \change{(e.g., \citep{long13,barfoot_tro14})}, where a rotational distribution can have a density of the form
\begin{equation}\label{eq:expdist}
p(\mbf{C}) \propto \exp\left( -\frac{1}{2} \ln(\mbf{C}\bar{\mbf{C}}^T)^{\vee^T} \mbs{\Sigma}^{-1} \ln(\mbf{C}\bar{\mbf{C}}^T)^{\vee} \right),
\end{equation}
where $\mbf{C}$ is a member of the matrix Lie group $SO(3)$, $\bar{\mbf{C}}$ is also a member of $SO(3)$ and represents the mean, and $\mbs{\Sigma}$ is an anisotropic matrix covariance.  We also have $\exp(\cdot)$ as the matrix exponential, $\ln(\cdot)$ the matrix logarithm, and $\vee$ a Lie algebra operator detailed a bit later in the paper.  Here we are essentially defining a Gaussian distribution in the vector space of the Lie algebra associated with $SO(3)$ and then mapping the uncertainty to the Lie group through the matrix exponential.  This allows for anisotropic distributions and the same approach can be easily extended to any matrix Lie group, such as the special Euclidean group $SE(3)$ that represents poses \change{(see, for example, \citep{long13, barfoot_tro14})}.  Our aim in this paper is to present some novel convex relaxations where rotational uncertainty is defined closer to this exponential coordinate model; to achieve this, we use the Cayley map, which is \change{very } close to the exponential map for small-to-moderate rotational uncertainty.  Our Cayley distributions will have the form
\begin{equation}\label{eq:caydist}
p(\mbf{C}) \propto \cay\left( -\frac{1}{2} \cay^{-1}(\mbf{C}\bar{\mbf{C}}^T)^{\vee^T} \mbs{\Sigma}^{-1} \cay^{-1}(\mbf{C}\bar{\mbf{C}}^T)^{\vee} \right),
\end{equation}
where $\cay(\cdot)$ is the Cayley map.  This also allows us to define our optimization problems directly on $SE(3)$ rather than $SO(3) \times \mathbb{R}^3$ when poses are involved.  

To our knowledge, the examination of global optimality for state estimation problems where rotational (and pose) uncertainty is defined in this way has not be explored previously in the literature. \change{Our novel contribution is therefore a family of specific convex relaxations of rotation and pose estimation problems formulated using the Cayley map (including redundant constraints needed to make them work in practice); this is important as it opens the door to providing certification for a broad class of state estimation problems used in practice.}

This paper is organized as follows.  In Section~\ref{sec:math} we review the relevant mathematical background including Lie groups, the Cayley map, and the convex relaxation procedure that we will employ.  Section~\ref{sec:averaging} presents the basic problems of averaging a number of noisy rotation or pose measurements.  In Section~\ref{sec:discretetime}, we expand the method to include discrete-time trajectory estimation of several poses based on individual and inter-pose measurements.  Section~\ref{sec:continuoustime} expands this to include so-called continuous-time trajectory estimation where we have a smoothing assumption on the trajectory and estimate both pose and twist at each state.  In each of Sections 3 to 5, we provide experimental results that demonstrate the viability of our convex relaxations to find globally optimal solutions.  Section~\ref{sec:conclusion} concludes the paper.  \change{Appendix~\ref{sec:distributions} discusses the similarity between distributions defined using the exponential and Cayley maps while } Appendix~\ref{sec:localsolvers} presents the baseline local solvers to which we compare our new global estimates.

\section{Mathematical Background}
\label{sec:math}

We begin by reviewing the relevant background concepts for the paper including Lie groups, the Cayley map, and convex relaxations of nonconvex optimization problems via Lagrangian duality.  

\subsection{Lie Groups for Rotations and Poses}

The {\em special orthogonal group}, representing rotations, is the set of valid rotation matrices:
\begin{equation}
\label{eq:SO3}
SO(3) = \left\{  \mbf{C} \in \mathbb{R}^{3\times3} \; | \; \mbf{C} \mbf{C}^T = \mbf{I}, \mbox{det}(\mbf{C}) = 1 \right\},
\end{equation}
where $\mbf{I}$ is the identity matrix.  
It is common to map a vector, $\mbs{\phi} \in \mathbb{R}^3$, to a rotation matrix, $\mbf{C}$, through the matrix exponential,
\begin{equation}
\mbf{C}(\mbs{\phi}) = \exp\left( \mbs{\phi}^\wdg \right),
\end{equation}
where $(\cdot)^\wdg$ is the skew-symmetric operator,
\begin{equation}
\mbs{\phi}^\wdg = \bbm \phi_1 \\ \phi_2 \\ \phi_3 \ebm^\wdg = \bbm 0 & -\phi_3 & \phi_2 \\ \phi_3 & 0 & -\phi_1 \\ -\phi_2 & \phi_1 & 0 \ebm,
\end{equation}
and $\mbs{\phi}  = \varphi \mbf{a} \in \mathbb{R}^3$, the product of the angle and unit axis of rotation.
The mapping is surjective-only, meaning every $\mbf{C}$ can be produced by many different values for $\mbs{\phi}$.

The {\em special Euclidean group}, representing \index{poses} poses (i.e., translation and rotation), is the set of valid  transformation matrices:
\begin{equation}
\label{eq:se3}
\scalebox{0.93}{
$SE(3) = \left\{  \mbf{T} = \bbm \mbf{C} & \mbf{r} \\ \;\,\mbf{0}^T & 1 \ebm \in \mathbb{R}^{4\times4} \; \Biggl| \; \mbf{C} \in SO(3), \, \mbf{r} \in \mathbb{R}^3 \right\}.$
}
\end{equation}
It is again common to map a vector, $\mbs{\xi} \in \mathbb{R}^6$, to a transformation matrix, $\mbf{T} \in SE(3)$, through the matrix exponential,
\begin{equation}
\mbf{T}(\mbs{\xi}) = \exp\left( \mbs{\xi}^\wdg \right),
\end{equation}
where we have overloaded the $\wdg$ operator as
\begin{equation}
\mbs{\xi}^\wdg = \bbm \mbs{\rho} \\ \mbs{\phi} \ebm^\wdg = \bbm \mbs{\phi}^\wdg & \mbs{\rho} \\ \mbf{0}^T & 0 \ebm.
\end{equation}
Notationally, we will use $\vee$ to mean the inverse operation of $\wdg$.
As is common practice \citep{barfoot_ser24}, we have broken the pose vector, $\mbs{\xi}$, into a translational component, $\mbs{\rho}$, and a rotational component, $\mbs{\phi}$.
The mapping is also surjective-only, meaning every $\mbf{T}$ can be produced by many different values for $\mbs{\xi}$.

Finally, the {\em adjoint} of a pose is given by
\begin{equation}\label{eq:se3adjointmap1}
\Tbig(\mbs{\xi}) = \mbox{Ad}\left( \mbf{T} \right) = \bbm \mbf{C} & \mbf{r}^\wdg \mbf{C} \\ \mbf{0} & \mbf{C} \ebm,
\end{equation}
which is now $6 \times 6$.  We will refer to the set of adjoints as $\mbox{Ad}(SE(3))$.  We can map a vector, $\mbs{\xi} \in \mathbb{R}^6$, to an adjoint transformation matrix again through the matrix exponential map:
\begin{equation}
\Tbig(\mbs{\xi}) = \exp\left( \mbs{\xi}^\Wdg \right),
\end{equation}
where
\begin{equation}
\mbs{\xi}^\Wdg = \bbm \mbs{\rho} \\ \mbs{\phi} \ebm^\Wdg = \bbm \mbs{\phi}^\wdg & \mbs{\rho}^\wdg \\ \mbf{0} & \mbs{\phi}^\wdg \ebm.
\end{equation}
Notationally, we will use $\uWdg$ to mean the inverse operation of $\Wdg$.
The mapping is again surjective-only, meaning every $\Tbig$ can be produced by many different values for $\mbs{\xi}$.  

\begin{figure*}[t]
\centering
\includegraphics[width=0.49\textwidth]{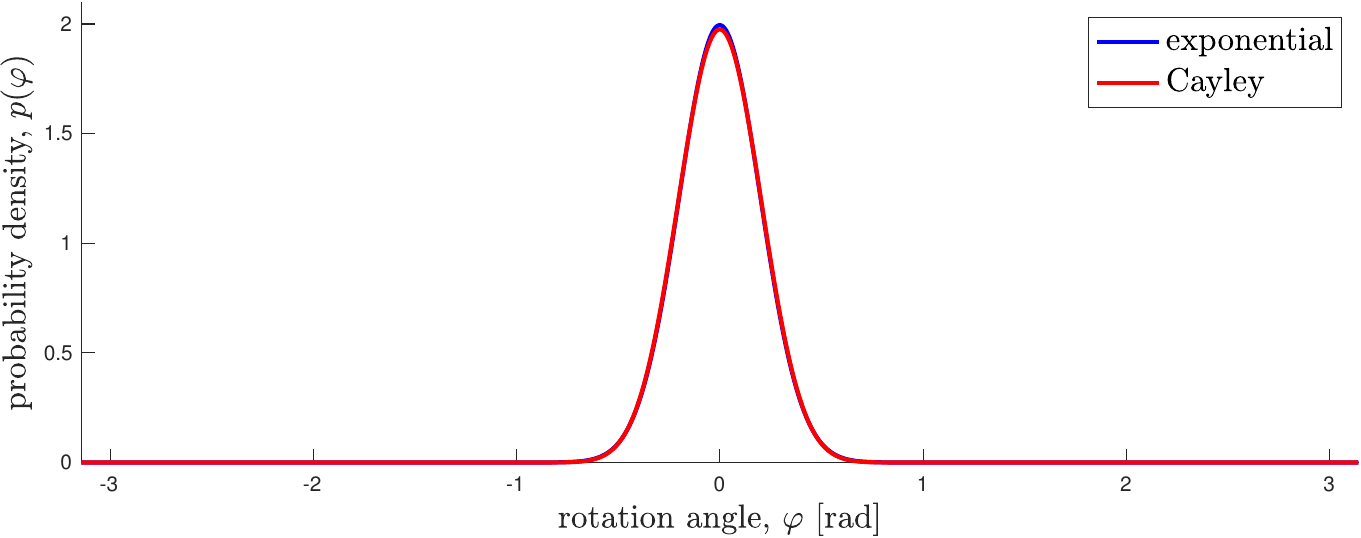} \hspace{0.05in} \includegraphics[width=0.49\textwidth]{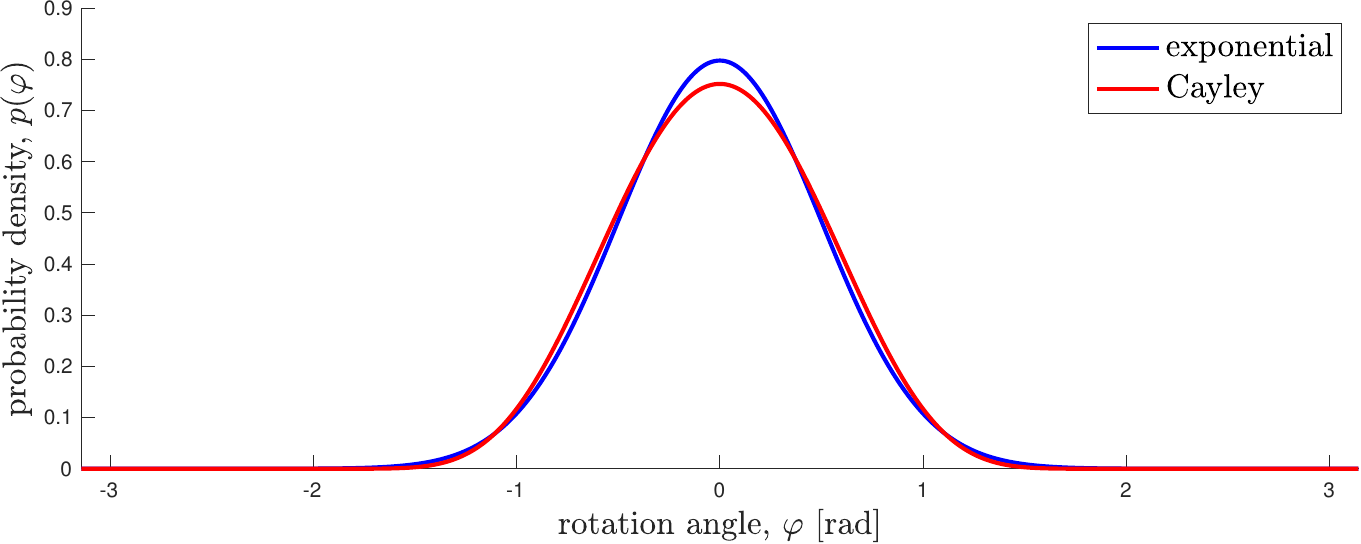}
\caption{\change{Comparison of uncertainty  on rotation angle, $\varphi$, for the exponential and Cayley maps, where the variances have been approximately matched (see Appendix~\ref{sec:distributions} for further discussion of how this was done).  (left) Standard deviation of rotational uncertainty is $\sigma = 0.2$ [rad].  (right) $\sigma = 0.5$ [rad].  The match is good in both cases with more divergence as rotational uncertainty increases.}}
\label{fig:maps}
\end{figure*}

\subsection{Cayley Map}

While the exponential map is the canonical way to map from a Lie algebra (vector space) to a Lie group, it is not the only possibility.  There are in fact infinitely many such vectorial mappings for $SO(3)$  \citep{bauchau03}, $SE(3)$ \citep{barfoot_rob22}, and $\mbox{Ad}(SE(3))$ \citep{bauchau03b,bauchau11}.  

In particular, it is well known that for the Cayley-Gibbs-Rodrigues parameterization of rotation we can write the rotation matrix in terms of the Cayley map, $\cay(\mbf{A}) = \left( \mbf{I} - \frac{1}{2}\mbf{A} \right)^{-1}  \left( \mbf{I} + \frac{1}{2} \mbf{A} \right)$, according to
 \citep{borri00,bauchau03}:
\beqn{}
\mbf{C}(\mbs{\phi}) & = & \cay(\mbs{\phi}^\wdg) = \left( \mbf{I} - \frac{1}{2}\mbs{\phi}^\wdg \right)^{-1}  \left( \mbf{I} + \frac{1}{2}\mbs{\phi}^\wdg \right), \nonumber \\  \\ \mbs{\phi} & = & \cay^{-1}(\mbf{C})^\vee = \left( 2 (\mbf{C} - \mbf{I}) (\mbf{C} + \mbf{I})^{-1} \right)^\vee, \qquad
\eeqn
for some $\mbs{\phi} = 2 \tan \frac{\varphi}{2} \mbf{a} \in \mathbb{R}^3$ with $\varphi$ the rotation angle and $\mbf{a}$ the unit axis.  \citet{borri00} and later \citet{selig07} demonstrated that the Cayley map can also be used to map pose vectors to $SE(3)$ according to
\beqn{se3cay}
\mbf{T}(\mbs{\xi}) & = & \cay(\mbs{\xi}^\wdg) = \left( \mbf{I} - \frac{1}{2}\mbs{\xi}^\wdg \right)^{-1}  \left( \mbf{I} + \frac{1}{2}\mbs{\xi}^\wdg \right), \nonumber \\ \\  \mbs{\xi} & = & \cay^{-1}(\mbf{T})^\vee = \left( 2 (\mbf{T} - \mbf{I}) (\mbf{T} + \mbf{I})^{-1} \right)^\vee, \qquad
\eeqn
for some $\mbs{\xi} \in \mathbb{R}^6$.  Although we will not need it, the Cayley map can be used to map pose vectors to $\mbox{Ad}(SE(3))$ according to
\beqn{se3Adcay}
\Tbig(\mbs{\xi}) & = & \cay(\mbs{\xi}^\Wdg) = \left( \mbf{I} - \frac{1}{2}\mbs{\xi}^\Wdg \right)^{-1}  \left( \mbf{I} + \frac{1}{2}\mbs{\xi}^\Wdg \right), \nonumber \\ \\  \mbs{\xi} & = & \cay^{-1}(\mbf{\Tbig})^\curlyvee = \left( 2 (\mbf{\Tbig} - \mbf{I}) (\mbf{\Tbig} + \mbf{I})^{-1} \right)^\uWdg. \qquad\;
\eeqn
\change{However}, \citet{selig07} demonstrates that starting from the same $\mbs{\xi}$ and applying~\eqref{eq:se3cay} and~\eqref{eq:se3Adcay} does not result in an equivalent transformation, i.e., $\Tbig(\mbs{\xi}) \neq \mbox{Ad}(\mbf{T}(\mbs{\xi}))$; the commutative property for adjoints does not hold.  Nevertheless, we shall not require this property here.

\change{Figure~\ref{fig:maps} provides examples comparing two rotational distributions derived from the exponential and Cayley maps that have approximately the same variance; we can see that even with quite large rotational uncertainty they match quite closely.  Appendix~\ref{sec:distributions} provides some further discussion on how closely these distributions can be made to match.  The key idea of the paper will be to replace instances of the exponential map with the Cayley map, which we will see is more amenable to producing polynomial optimization problems, a key prerequisite to our route to global optimality.}

The Cayley map has been used in the past for rotation\change{, pose, and trajectory }estimation \citep{mortari07,majji11,junkins11,alismail_continuous_2014,wong16,wong18,qian20,barfoot_rob22}, typically to parameterize rotations or poses thereby creating a simpler unconstrained quadratic optimization problem. The drawback of these approaches is that they are still subject to singularities and local minima.  The Cayley map has also been used to achieve global optimality in the perspective-$n$-point (PnP) problem \citep{nakano15,wang22}; we take a quite different approach, however, through the use of convex relaxations. 
\change{Additionally, the Cayley map has found application in parametrizing lines in structure from motion, as an unconstrained alternative to parametrizations such as Pl\"{u}cker coordinates~\citep{zhang_structure_2014}.}

\change{The Cayley map has also been employed in areas of robotics other than estimation. In~\citep{kobilarov_discrete_2011,kobilarov_discrete_2014}, for example, the authors suggest using the Cayley map for rotation parametrization in the context of optimal control of mechanical systems on Lie groups. The authors observe that, compared with the exponential map, the Cayley map is computationally more efficient because of its simpler structure, in particular as it circumvents trigonometric functions. It is also noted that the Cayley map has no singularities in its gradients, which is of advantage for the numerical stability of commonly used solvers~\citep{kobilarov_discrete_2011}. In~\citep{solo_numerical_2019}, the Cayley map is employed for simulating stochastic differential equations that evolve on Stiefel manifolds, which is subsequently used in~\citep{wang_lie_2020} for a novel particle filter variant.  It is possible that our global optimality approach to using the Cayley map could be employed within some of these applications, but we leave this investigation for future work.}

\subsection{Convex Relaxations}
\label{sec:convexrelaxations}

We next summarize the key optimization tools that we will use.  \citet{boyd04} provide the appropriate background.  Suppose that we have a nonconvex optimization problem of the form
\begin{equation}
\begin{tabular}{rl}
$\min$ & $f(\mbf{z})$ \\
w.r.t. & $\mbf{z}$ \\
s.t. & $g_i(\mbf{z}) = 0 \quad (\forall i)$
\end{tabular}.
\end{equation}
We attempt to introduce appropriate nonlinear substitution variables, $\mbf{x}$, to replace $\mbf{z}$ so that both the objective and the constraints can be written in a standard, homogeneous, quadratic form:
\begin{equation}
\begin{tabular}{rl}
$\min$ & $\mbf{x}^T \mbf{Q} \mbf{x}$ \\
w.r.t. & $\mbf{x}$ \\
s.t. & $\mbf{x}^T \mbf{A}_0 \mbf{x} = 1$ \\
& $\mbf{x}^T \mbf{A}_i \mbf{x} = 0 \quad (\forall i \neq 0)$
\end{tabular}.
\end{equation}
This problem is a \acf{QCQP}, which is still nonconvex and typically of higher dimension than the original problem, but possesses more exploitable structure.  Next, by defining $\mbf{X} = \mbf{x} \mbf{x}^T$, we rewrite this problem exactly as
\begin{equation}
\begin{tabular}{rl}
$\min$ & $\mbox{tr}\left(\mbf{Q}\mbf{X}\right)$ \\
w.r.t. & $\mbf{X}$ \\
s.t. & $\mbf{X} \succeq 0$ \\
& $\mbox{rank}(\mbf{X}) = 1$ \\
&  $\mbox{tr}\left(\mbf{A}_0 \mbf{X} \right) = 1$ \\
& $\mbox{tr}\left(\mbf{A}_i \mbf{X} \right) = 0 \quad (\forall i \neq 0) $ 
\end{tabular}.
\end{equation}
Finally, if we drop the $\mbox{rank}(\mbf{X}) = 1$ constraint, we have a {\em convex relaxation} of the problem in the form of a \acf{SDP}:
\begin{equation}\label{eq:sdp1}
\begin{tabular}{rl}
$\min$ & $\mbox{tr}\left(\mbf{Q}\mbf{X}\right)$ \\
w.r.t. & $\mbf{X}$ \\
s.t. & $\mbf{X} \succeq 0$ \\
& $\mbox{tr}\left(\mbf{A}_0 \mbf{X} \right) = 1$ \\
& $\mbox{tr}\left(\mbf{A}_i \mbf{X} \right) = 0 \quad (\forall i \neq 0) $ 
\end{tabular}.
\end{equation}
This is known as {\em Shor's relaxation} \citep{shor87}.  If the solution to this problem happens to result in $\mbox{rank}(\mbf{X}) = 1$, then we have \change{an a posteriori}\footnote{\change{While $\mbox{rank}(\mbf{X}) = 1$ provides an a posteriori guarantee that our solution is globally optimal, we cannot a priori guarantee that we will get a rank-one solution for a given set of measurements.  Measurement noise, for example, can have a big impact on tightness and so we will later use empirical studies to a priori gauge how likely we are to find rank-one solutions at different measurement noise levels.}} guarantee that we also have a global solution to the original problem, $\mbf{x}$.  Since \acp{SDP} are convex problems, we can attempt to use standard solvers, such as interior-point methods, to solve them numerically.  While there are tractability issues to be addressed to scale up to very large problem instances, we will see that for the problems in this paper, this approach is viable for nontrivial problem sizes.

Unfortunately, for most problems in this paper, Shor's relaxation is not {\em tight} out of the box.  In this paper, tightness means $\mbox{rank}(\mbf{X}) = 1$ (and therefore that the optimal cost matches that of the original problem).  However, there is still a way forward.  We can attempt to introduce additional so-called 
{\em redundant constraints} to tighten up the relaxation. These constraints do not affect the feasible set of the original optimization problem, but they do reduce the feasible set of the \ac{SDP} in order to tighten it.

\begin{figure*}[t]
\centering
\includegraphics[width=0.47\textwidth]{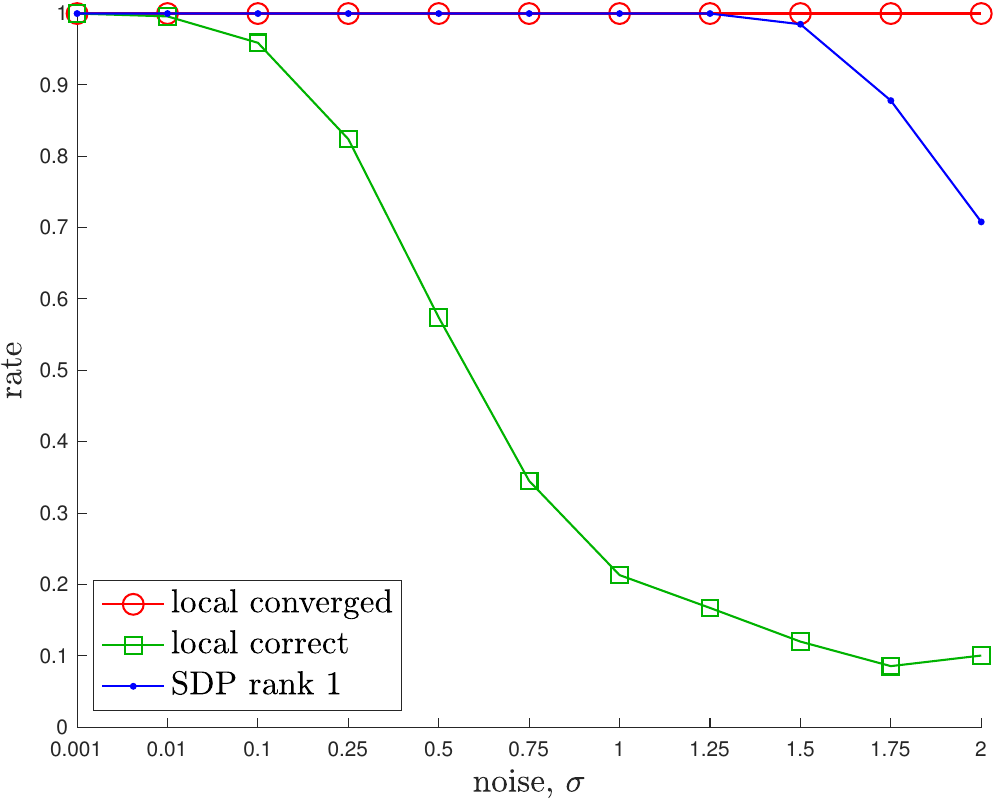}\hspace{0.2in}
\includegraphics[width=0.47\textwidth]{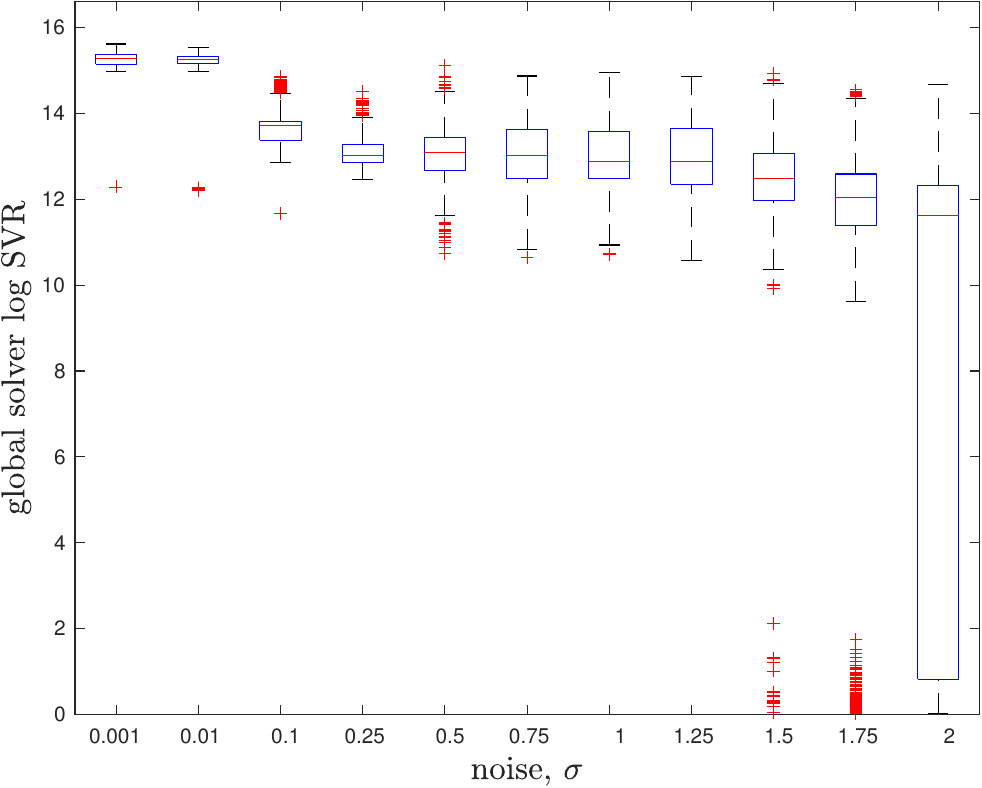} 
\caption{{\em Rotation Averaging:} A quantitative evaluation of the tightness of the rotation averaging problem with increasing measurement noise level, $\sigma$.  At each noise level, we conducted $1000$ trials of averaging $10$ noisy rotations. \change{(left) We see that the local solver (randomly initialized) finds the global minimum with decreasing frequency (green) as the measurement noise is increased, while the \ac{SDP} solver (blue) successfully produces rank-1 solutions (we consider log \ac{SVR} of at least $5$ to be rank 1) to much higher noise levels.  For completeness, we also show how frequently the local solver converges to any minimum (red). } (right) Boxplots\protect\footnotemark  of the log \ac{SVR} of the \ac{SDP} solution show that the global solution remains highly rank 1 over a wide range of measurement noise values.}
\label{fig:cayley_rotav2_noise}
\end{figure*}

\change{The technique of adding redundant constraints to improve the tightness of a given SDP relaxation has been known for some time in the optimization literature \citep{anstreicher00,nesterovSemidefiniteProgrammingRelaxations2000}. More recently, there have been several cases in which redundant constraints have been used in the robotics \citep{yang20,yang22,giamou19,wangCertifiablyOptimalMutual2022} and machine vision \citep{briales17b,brialesCertifiablyGloballyOptimal2018,garcia-salgueroTighterRelaxationRelative2022,kezurerTightRelaxationQuadratic2015} literature. With redundant constraints, our } problem becomes 
\begin{equation}\label{eq:sdp2}
\begin{tabular}{rl}
$\min$ & $\mbox{tr}\left(\mbf{Q}\mbf{X}\right)$ \\
w.r.t. & $\mbf{X}$ \\
s.t. & $\mbf{X} \succeq 0$ \\
& $\mbox{tr}\left(\mbf{A}_0 \mbf{X} \right) = 1$ \\
& $\mbox{tr}\left(\mbf{A}_i \mbf{X} \right) = 0 \quad (\forall i \neq 0) $ \\
(redundant) & $\mbox{tr}\left(\mbf{B}_j \mbf{X} \right) = 0 \quad (\forall j \neq 0) $ 
\end{tabular},
\end{equation}
where the $\mbf{B}_j$ encapsulate these additional redundant constraints.  In theory, \citet{lasserre01} tells us how to tighten our \ac{SDP}, if possible, by adding a progression of variables and constraints, but adding too many constraints can be computationally expensive and in practice not necessary for tightness.\footnote{\change{The interested reader is directed to \citep{henrionMomentSOSHierarchyLectures2020} and \citep{lasserreMomentsPositivePolynomials2010} for a more in depth treatment on the {\em moment/sum-of-squares hierarchy.}}}  On the other hand, devising a sufficient set of constraints can be challenging by trial and error.  In our concurrent work, we have been developing a tool to automatically find such constraints \citep{duembgen_tro23}, which we used to help identify some of the constraints reported in this paper.  In all of the pose estimation problems to follow, we do require redundant constraints and we will be explicit in enumerating ones that in practice result in tight \ac{SDP} relaxations of our problems.  This could be viewed as the core contribution of the paper.

\footnotetext{\change{On  each box, the central mark is the median, the edges of the box are the
    25th and 75th percentiles, the whiskers extend to the most extreme
    datapoints the algorithm considers to be not outliers, and the outliers
    are plotted individually (as red + symbols).}}

\begin{figure}[b]
\centering
\includegraphics[width=0.2\textwidth]{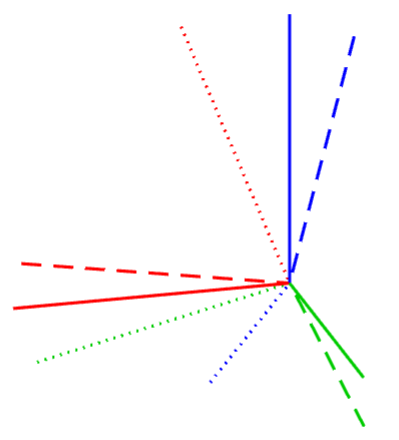}
\caption{{\em Rotation Averaging:} An example of noisy rotation averaging where the randomly initialized local solver (dotted) becomes trapped in a poor local minimum while the global solver (dashed) finds the correct global solution, which is closer to the groundtruth rotation (solid).}
\label{fig:cayley_rotav2}
\end{figure}

\section{Averaging}
\label{sec:averaging}

We will build up our optimization problems gradually starting with simply `averaging' several noisy estimates of rotation or pose.  

\subsection{Rotation Averaging}

In order to average $M$ rotations, we could set up an optimization problem as
\begin{equation}\label{eq:rotavprob}
\begin{tabular}{rl}
$\min$ & $\sum_{m=1}^M \ln\left(\mbf{C}\tilde{\mbf{C}}_m^T \right)^{\vee^T} \! \mbf{W}_m \, \ln\left(\mbf{C}\tilde{\mbf{C}}_m^T \right)^\vee$ \\
w.r.t. & $\mbf{C}$ \\
s.t. & $\mbf{C} \in SO(3)$
\end{tabular},
\end{equation}
where $\tilde{\mbf{C}}_m \in SO(3)$ are the noisy rotations to be averaged and $\mbf{W}_m$ is a matrix weight.  This type of cost function is used frequently in rotational estimation problems \citep{barfoot_ser24} and can represent a maximum-likelihood problem when the generative model for the noisy measurements is of the form
\begin{equation}
\tilde{\mbf{C}}_m = \exp\left( \mbs{\phi}_m^\wdg \right) \mbf{C}, \quad \mbs{\phi}_m \sim \mathcal{N}(\mbf{0}, \mbf{W}_m^{-1}).
\end{equation}
Alternatively, we can view our cost as the negative log-likelihood of the joint distribution of the measurements if each obeys~\eqref{eq:expdist}.  The trouble is that the matrix exponential and logarithm are difficult expressions to manipulate into the \ac{QCQP} form we seek. 

\begin{figure*}[t]
\centering
\includegraphics[width=\textwidth]{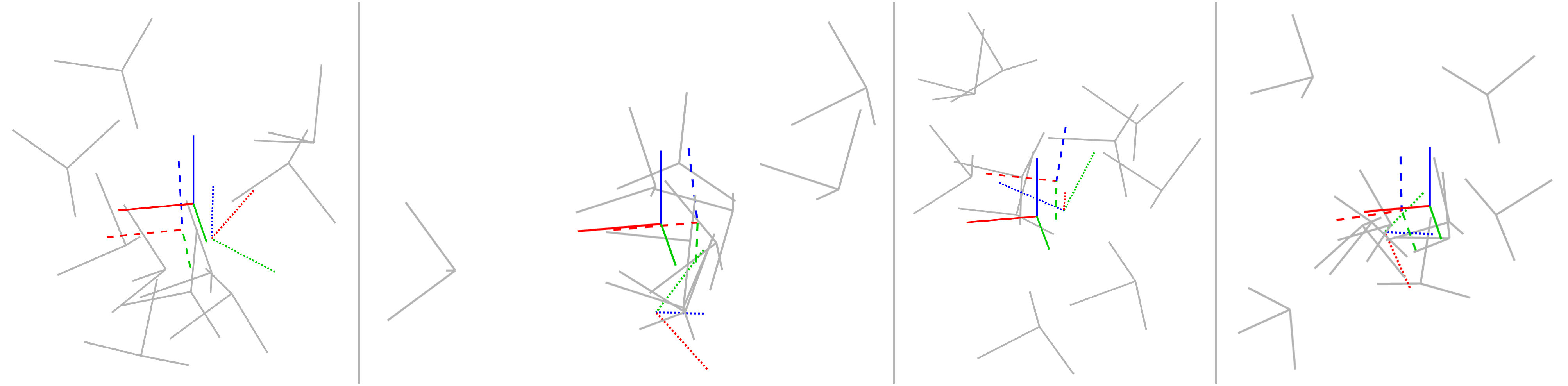}
\caption{{\em Pose Averaging:} Four examples of noisy pose averaging where the randomly initialized local solver (dotted) becomes trapped in a poor local minimum while the global solver (dashed) finds the correct solution, which is closer to the groundtruth pose (solid).  The noisy pose measurements being averaged are shown in grey.}
\label{fig:cayley_poseav2}
\end{figure*}

This is where the main insight of the paper comes in.  We can substitute the Cayley map for the exponential map without too much effect on the stated problem \change{(see Figure~\ref{fig:maps}).  With this substitution, }our generative model for noisy rotations becomes
\begin{equation}
\tilde{\mbf{C}}_m = \cay\left( \mbs{\phi}_m^\wdg \right) \mbf{C}, \quad \mbs{\phi}_m \sim \mathcal{N}(\mbf{0}, \mbf{W}_m^{-1}),
\end{equation}
and so our optimization problem can be restated as
\begin{equation}
\begin{tabular}{rl}
$\min$ & $\sum_{m=1}^M \cay^{-1}\left(\mbf{C}\tilde{\mbf{C}}_m^T \right)^{\vee^T} \! \mbf{W}_m \, \cay^{-1}\left(\mbf{C}\tilde{\mbf{C}}_m^T \right)^\vee$ \\
w.r.t. & $\mbf{C}$ \\
s.t. & $\mbf{C} \in SO(3)$
\end{tabular}.
\end{equation}
Now our cost represents the negative log-likelihood of the joint distribution of the measurements, assuming each obeys~\eqref{eq:caydist}.
Turning this into a \ac{QCQP} is then fairly easy:
\begin{equation}\label{eq:rotav}
\begin{tabular}{rl}
$\min$ & $\sum_{m=1}^M \mbs{\phi}_m^T \mbf{W}_m \mbs{\phi}_m$ \\
w.r.t. & $\mbf{c}_1, \mbf{c}_2, \mbf{c}_3, \mbs{\phi}_m \quad (\forall m)$ \\
s.t. & $\mbf{c}_i^T\mbf{c}_j = \delta_{ij} \quad (\forall i,j)$ \\
& $\left(\mbf{I} - \frac{1}{2} \mbs{\phi}_m^\wdg \right) \mbf{c}_i = \left(\mbf{I} + \frac{1}{2} \mbs{\phi}_m^\wdg \right) \tilde{\mbf{c}}_{m,i} \quad (\forall i, m)$
\end{tabular},
\end{equation}
where $\delta_{ij}$ is the Kronecker delta and
\begin{equation}
\mbf{C} = \bbm \mbf{c}_1 & \mbf{c}_2 & \mbf{c}_3 \ebm, \qquad \tilde{\mbf{C}}_m = \bbm \tilde{\mbf{c}}_{m,1} & \tilde{\mbf{c}}_{m,2} & \tilde{\mbf{c}}_{m,3} \ebm.  
\end{equation}
We have essentially introduced variables, $\mbs{\phi}_m$, for the residual errors of each term in the cost and used these to connect $\mbf{C}$ to each $\tilde{\mbf{C}}_m$ through the Cayley map; by bringing the inverse factor of the Cayley map to the other side, this becomes a quadratic constraint.  Thus we have both a quadratic cost and quadratic constraints and hence a \ac{QCQP}.  The dimension of the problem is now higher since we must now optimize over $\mbf{C}$ and all the $\mbs{\phi}_m$; however, we can follow the approach of Section~\ref{sec:convexrelaxations} to produce a \ac{SDP} relaxation of the problem.  Note, we have quietly dropped the $\det(\mbf{C}) = 1$ constraint on the rotation and will simply check it at the end\footnote{This is also a relaxation of the problem, but one that has been shown to often be tight in practice \citep{rosen19}.}; our optimization then only guarantees $\mbf{C} \in O(3)$ not $SO(3)$.  We leave the details of manipulating~\eqref{eq:rotav} into the standard form of~\eqref{eq:sdp1} to the reader.  We did not find any redundant constraints were necessary to tighten this relaxation; the \ac{SDP} solution produced remains rank 1 in practice for reasonably high amounts of noise.  

The details of a baseline local solver can be found in the appendix.  For the global (\ac{SDP}) solver we used \texttt{cvx} in Matlab with \texttt{mosek} \citep{mosek}.  The solution costs of the global and local solvers agree to high precision\footnote{\change{For this and all subsequent experiments, the local solver was verified to be converged in each case to a step size change of less than $10^{-6}$ while the global solver reported its own solution quality and used the \texttt{cvx\_precision high} setting (see the \href{https://cvxr.com/cvx/doc/solver.html}{cvx documentation)}.}} if a good initial guess is given to the local solver.   Figure~\ref{fig:cayley_rotav2} provides an example where the local solver converges from a poor initial guess to a local minimum, while the global solver finds the optimal solution near the groundtruth.  Figure~\ref{fig:cayley_rotav2_noise} provides a quantitative study of the tightness of the \ac{SDP} solution with increasing measurement noise; we selected the measurement covariance as $\mbf{W}_m^{-1} = \sigma^2 \mbf{I}$, with $\sigma$ increasing.  To gauge numerically whether the \ac{SDP} solution, $\mbf{X}$, is rank 1, we define the {\em logarithmic \acf{SVR}} as the base-10 logarithm of the ratio of the largest to second-largest singular values of $\mbf{X}$; we consider a log \ac{SVR} \change{of at least } $5$ to represent rank 1.  We see there is a large range for the noise over which the local solver can become trapped in a local minimum while the global solver remains rank 1.  With $M = 10$ rotations to be averaged, the local solver \change{took on average $0.0012s$ while the \ac{SDP} solver took on average $0.3486s$.}  

\begin{figure*}[t]
\centering
\includegraphics[width=0.47\textwidth]{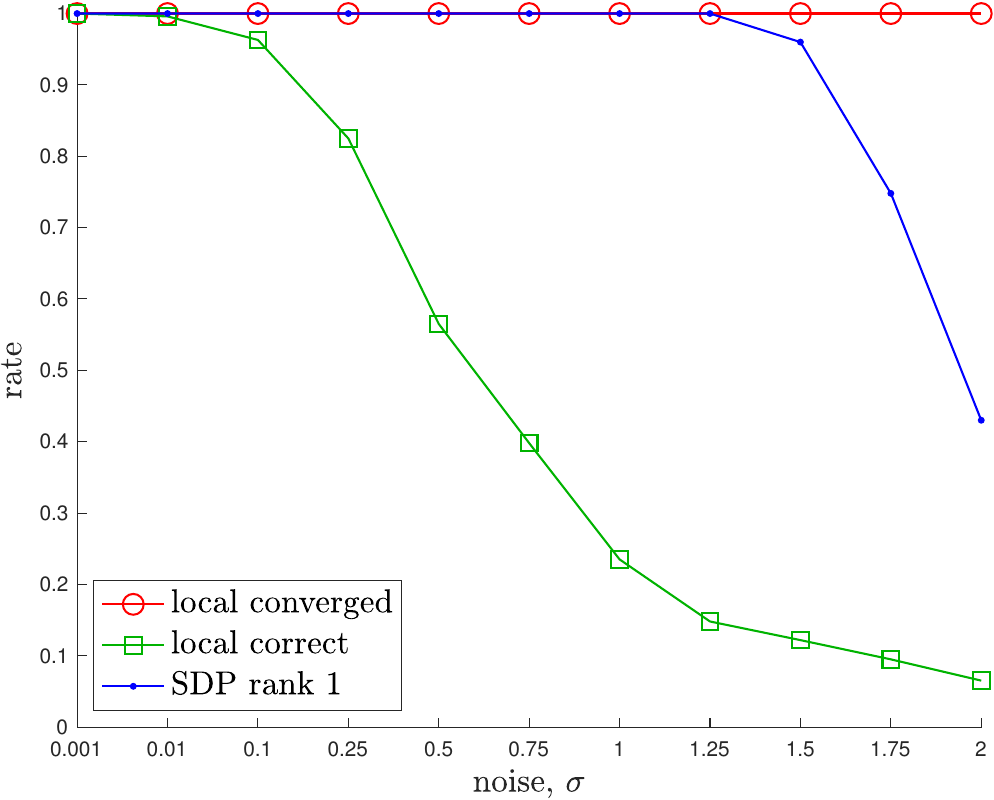}\hspace{0.2in}
\includegraphics[width=0.47\textwidth]{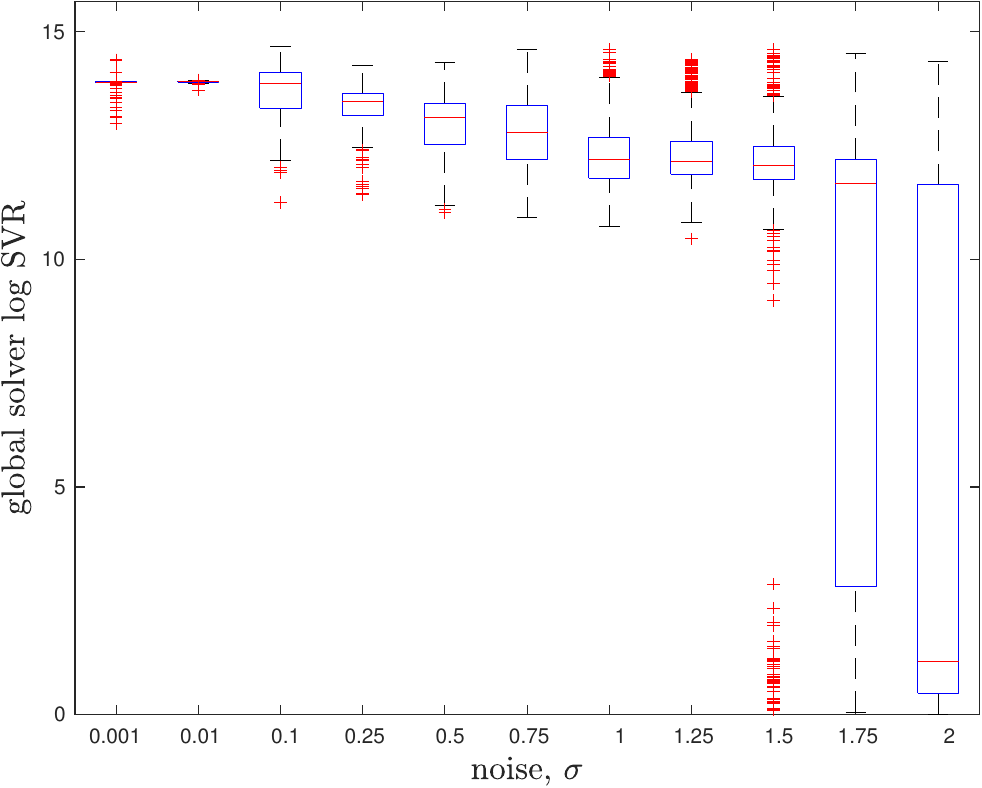} 
\caption{{\em Pose Averaging:} A quantitative evaluation of the tightness of the pose averaging problem with increasing measurement noise level, $\sigma$.  At each noise, we conducted $1000$ trials of averaging $10$ noisy poses.  \change{(left) We see that the local solver (randomly initialized) finds the global minimum with decreasing frequency (green) as the measurement noise is increased, while the \ac{SDP} solver (blue) successfully produces rank-1 solutions (we consider log \ac{SVR} of at least $5$ to be rank 1) to much higher noise levels.  For completeness, we also show how frequently the local solver converges to any minimum (red). } (right) Boxplots of the log \ac{SVR} of the \ac{SDP} solution show that the global solution remains highly rank 1 over a wide range of measurement noise values.}
\label{fig:cayley_poseav3_noise}
\end{figure*}

\subsection{Pose Averaging}

Pose averaging follows a very similar approach to the previous section.  An optimization problem based on the Cayley map can be stated as
\begin{equation}\label{eq:poseavprob}
\begin{tabular}{rl}
$\min$ & $\sum_{m=1}^M \cay^{-1}\left(\mbf{T}\tilde{\mbf{T}}_m^{-1} \right)^{\vee^T} \! \mbf{W}_m \, \cay^{-1}\left(\mbf{T}\tilde{\mbf{T}}_m^{-1} \right)^\vee$ \\
w.r.t. & $\mbf{T}$ \\
s.t. & $\mbf{T} \in SE(3)$
\end{tabular},
\end{equation}
where $\tilde{\mbf{T}}_m$ are noisy pose measurements with matrix weights, $\mbf{W}_m$.

\begin{figure*}[t]
\centering
\includegraphics[width=\textwidth]{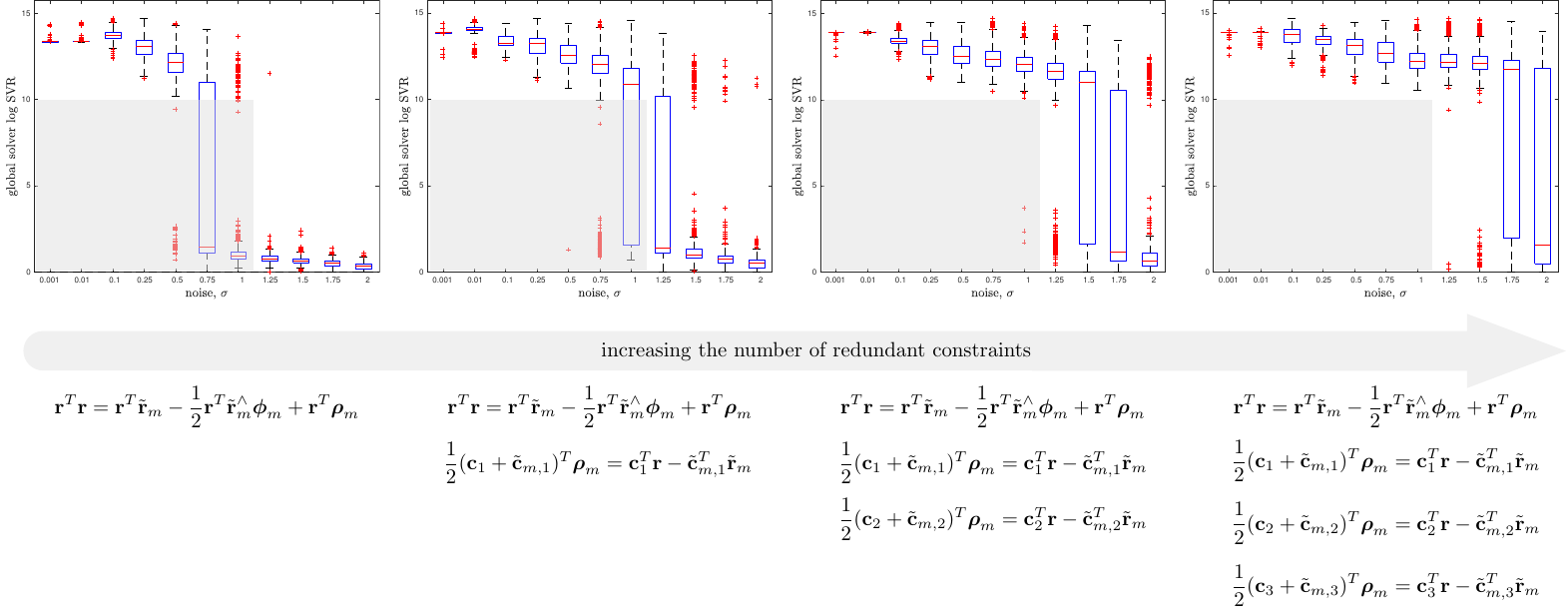} 
\caption{\change{{\em Pose Averaging Ablation Study:} Here we show the effect on \ac{SDP} tightness of varying the number of redundant constraints in the pose averaging problem.  The rightmost column shows our full set of recommended redundant constraints with the light-grey box indicating the region of measurement noise for which our problem can be deemed tight.  The same grey box is shown in the other columns for reference, indicating that including fewer redundant constraints results in a lower level of noise for which we can keep the solution tight.}}
\label{fig:ablation}
\end{figure*}

We convert the residual pose errors, $\mbs{\xi}_m = \cay^{-1}\left(\mbf{T}\tilde{\mbf{T}}_m^{-1} \right)^\vee$, to variables and now our optimization can be stated as
\begin{equation}
\begin{tabular}{rl}
$\min$ & $\sum_{m=1}^M \mbs{\xi}_m^T \mbf{W}_m \mbs{\xi}_m$ \\
w.r.t. & $\mbf{T}, \mbs{\xi}_m \quad (\forall m)$ \\
s.t. & $\mbf{C}^T\mbf{C} = \mbf{I}$ \\
& $\left(\mbf{I} - \frac{1}{2} \mbs{\xi}_m^\wdg \right) \mbf{T} = \left(\mbf{I} + \frac{1}{2} \mbs{\xi}_m^\wdg \right) \tilde{\mbf{T}}_m \quad (\forall m)$
\end{tabular},
\end{equation}
where we have again dropped the $\det(\mbf{C}) = 1$ constraint.
Since the bottom row of a transformation matrix is constant, we can parameterize it in the following way
\beqn{}
\mbf{T} & = & \bbm \mbf{C} & \mbf{r} \\ \mbf{0}^T & 1 \ebm = \bbm \mbf{c}_1 & \mbf{c}_2 & \mbf{c}_3 & \mbf{r} \\ 0 & 0 & 0 & 1 \ebm, \\ \tilde{\mbf{T}}_m & = & \bbm \tilde{\mbf{C}}_m & \mbf{r}_m \\ \mbf{0}^T & 1 \ebm = \bbm \tilde{\mbf{c}}_{m,1} & \tilde{\mbf{c}}_{m,2} & \tilde{\mbf{c}}_{m,2} & \tilde{\mbf{r}}_m \\ 0 & 0 & 0 & 1 \ebm, \nonumber \\ \\ \mbs{\xi}_m & = & \bbm \mbs{\rho}_m \\ \mbs{\phi}_m \ebm,
\eeqn
and then rewrite the optimization problem using the reduced set of variables as
\begin{equation}\label{eq:poseav}
\begin{tabular}{rl}
$\min$ & $\sum_{m=1}^M \mbs{\xi}_m^T \mbf{W}_m \mbs{\xi}_m$ \\
w.r.t. & $\mbf{c}_i, \mbf{r}, \mbs{\rho}_m, \mbs{\phi}_m \quad (\forall i, m)$ \\
s.t. & $\mbf{c}_i^T\mbf{c}_j = \delta_{ij} \quad (\forall i,j)$ \\
& $(\mbf{I} -  \frac{1}{2}\mbs{\phi}_m^\wdg) \mbf{c}_i = (\mbf{I} +  \frac{1}{2}\mbs{\phi}_m^\wdg) \tilde{\mbf{c}}_{m,i} \quad (\forall i,m)$ \\
& $(\mbf{I} -  \frac{1}{2}\mbs{\phi}_m^\wdg) \mbf{r} = (\mbf{I} +  \frac{1}{2}\mbs{\phi}_m^\wdg) \tilde{\mbf{r}}_m + \mbs{\rho}_m \quad (\forall m)$
\end{tabular}.
\end{equation}
This is now a \ac{QCQP}, but unfortunately when we relax to a \ac{SDP}, it is not always tight \change{even for low noise levels}.  We found that introducing specific redundant constraints for each $m$ tightens the problem nicely \change{for practical noise levels}.  One such useful constraint can be found by premultiplying the last constraint in~\eqref{eq:poseav} by $\mbf{r}^T$ whereupon
\begin{equation}
\mbf{r}^T\mbf{r} -  \frac{1}{2} \underbrace{\mbf{r}^T\mbs{\phi}_m^\wdg \mbf{r}}_{\mbf{0}} = \mbf{r}^T \tilde{\mbf{r}}_m  -  \frac{1}{2} \mbf{r}^T  \tilde{\mbf{r}}_m^\wdg \mbs{\phi}_m + \mbf{r}^T\mbs{\rho}_m.
\end{equation}
The key is that the second cubic term vanishes, leaving a new quadratic constraint that it is not simply a trivial linear combination of the existing constraints \citep{yang22}. However, this constraint is redundant because it does not restrict the original feasible set at all.  In the lifted \ac{SDP} space it serves to restrict the feasible set and ultimately tighten the relaxation.  
Another useful redundant constraint can be formed by combining the last two of~\eqref{eq:poseav}; the second last can be written as
\begin{equation}
\frac{1}{2} ( \mbf{c}_i + \tilde{\mbf{c}}_{m,i} )^T \mbs{\phi}_m^\wdg  = - \left(\mbf{c}_i - \tilde{\mbf{c}}_{m,i} \right)^T,
\end{equation}
while the last can be premultiplied by $( \mbf{c}_i + \tilde{\mbf{c}}_{m,i} )^T$ and written as
\begin{multline}
( \mbf{c}_i + \tilde{\mbf{c}}_{m,i} )^T ( \mbf{r} - \tilde{\mbf{r}}_m ) = \underbrace{\frac{1}{2} ( \mbf{c}_i + \tilde{\mbf{c}}_{m,i} )^T \mbs{\phi}_m^\wdg}_{-(\mbf{c}_i - \tilde{\mbf{c}}_{m,i})^T} ( \mbf{r} + \tilde{\mbf{r}}_m ) \\ + ( \mbf{c}_i + \tilde{\mbf{c}}_{m,i} )^T \mbs{\rho}_m.
\end{multline}
After performing the indicated substitution, this becomes
\begin{equation}
\frac{1}{2} ( \mbf{c}_i + \tilde{\mbf{c}}_{m,i} )^T \mbs{\rho}_m = \mbf{c}_i^T \mbf{r} - \tilde{\mbf{c}}_{m,i}^T \tilde{\mbf{r}}_m,
\end{equation}
which is once again a quadratic constraint.

Summarizing, the following \ac{QCQP} offers a reasonably tight \ac{SDP} relaxation in practice:
\begin{equation}\label{eq:poseav2}
\begin{tabular}{rl}
$\min$ & $\sum_{m=1}^M \mbs{\xi}_m^T \mbf{W}_m \mbs{\xi}_m$ \\
w.r.t. & $\mbf{c}_i, \mbf{r}, \mbs{\rho}_m, \mbs{\phi}_m \quad (\forall i, m)$ \\
s.t. & $\mbf{c}_i^T\mbf{c}_j = \delta_{ij} \quad (\forall i,j)$ \\
& $(\mbf{I} -  \frac{1}{2}\mbs{\phi}_m^\wdg) \mbf{c}_i = (\mbf{I} -  \frac{1}{2}\mbs{\phi}_m^\wdg) \tilde{\mbf{c}}_{m,i} \quad (\forall i,m)$ \\
& $(\mbf{I} -  \frac{1}{2}\mbs{\phi}_m^\wdg) \mbf{r} = (\mbf{I} +  \frac{1}{2}\mbs{\phi}_m^\wdg) \tilde{\mbf{r}}_m + \mbs{\rho}_m \quad (\forall m)$ \\
(red.) & $ \frac{1}{2} ( \mbf{c}_i + \tilde{\mbf{c}}_{m,i} )^T \mbs{\rho}_m = \mbf{c}_i^T \mbf{r} - \tilde{\mbf{c}}_{m,i}^T \tilde{\mbf{r}}_m \quad (\forall i,m)$ \\
& $\mbf{r}^T\mbf{r}  = \mbf{r}^T \tilde{\mbf{r}}_m  -  \frac{1}{2} \mbf{r}^T  \tilde{\mbf{r}}_m^\wdg \mbs{\phi}_m + \mbf{r}^T\mbs{\rho}_m \quad (\forall m)$
\end{tabular}.
\end{equation}
We leave it to the reader to manipulate this into the standard form of~\eqref{eq:sdp2}.

\begin{figure*}[t]
\centering
\includegraphics[width=0.53\textwidth]{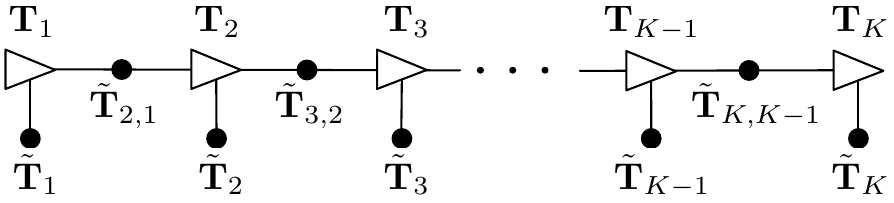}
\caption{{\em Discrete-time Trajectory Estimation:}  Factor graph representation of the discrete-time estimation problem.  Each black dot represents one of the error terms in the cost function of~\eqref{eq:trajestprob}.}
\label{fig:trajest}
\vspace*{-0.1in}
\end{figure*}

The appendix again provides a baseline local solver for this problem.  For the global (SDP) solver we used \texttt{cvx} in Matlab with \texttt{mosek} \citep{mosek}.  The solution costs of the global and local solvers agree to high precision if a good initial guess is given to the local solver.  Figure~\ref{fig:cayley_poseav2} provides examples of the local solver becoming trapped in poor local minima while the global solver converges to the correct minima near the groundtruth.  Figure~\ref{fig:cayley_poseav3_noise} provides a quantitative study of the tightness of the \ac{SDP} solution with increasing measurement noise; we selected the measurement covariance as $\mbf{W}_m^{-1} = \sigma^2 \mbf{I}$, with $\sigma$ increasing.  We again see there is a large range for the noise over which the local solver can become trapped in a local minimum while the global solver remains tight.  With $M = 10$ poses to be averaged, the local solver took on average \change{$0.0064s$ while the SDP solver took on average $0.5944s$.}

\change{To justify the need for the redundant constraints, we conducted an ablation study (see Figure~\ref{fig:ablation}) wherein we varied the number of redundant constraints.  The study shows that with more redundant constraints, we can tolerate a higher level of measurement noise while keeping the \ac{SDP} tight.  We always included the last redundant constraint in~\eqref{eq:poseav2} as this enforces that the search space for the \ac{SDP} remains compact\footnote{\change{Compactness in this context is related to the Archimedean property of the feasible set, which is a required condition when tightening a relaxation using Lasserre's hierarchy \citep{lasserreMomentsPositivePolynomials2010}.}} and it is therefore well posed.  In the interest of space, we forgo similar ablation studies for the subsequent problems (discrete-time and continuous-time trajectory estimation), which reuse the pose averaging redundant constraints and then build on top of them.  The studies are similar in that the more redundant constraints we add, the larger the noise region for which we can a priori predict that we will achieve rank-1 \ac{SDP} solutions.}

\section{Discrete-Time Trajectory Estimation}
\label{sec:discretetime}

Our next problem is to consider estimation of a trajectory of $K$ poses, $\mbf{T}_k$, where we have noisy measurements of each pose, $\tilde{\mbf{T}}_k$, as well as noisy relative measurements, $\tilde{\mbf{T}}_{k+1,k}$, from one pose to the next.  The optimization problem that we want to solve is
\begin{equation}\label{eq:trajestprob}
\begin{tabular}{rl}
$\min$ & $\sum_{k=1}^K \cay^{-1}\left(\mbf{T}_k\tilde{\mbf{T}}_k^{-1} \right)^{\vee^T} \! \mbf{W}_k \, \cay^{-1}\left(\mbf{T}_k\tilde{\mbf{T}}_k^{-1} \right)^\vee$ \\ & \quad $+ \; \sum_{k=1}^{K-1} \cay^{-1}\left(\mbf{T}_{k+1}\mbf{T}_k^{-1}\tilde{\mbf{T}}_{k+1,k}^{-1} \right)^{\vee^T}$ \\ & \quad $\quad \times \; \mbf{W}_{k+1,k} \, \cay^{-1}\left(\mbf{T}_{k+1} \mbf{T}_k^{-1}\tilde{\mbf{T}}_{k+1,k}^{-1} \right)^\vee$ \\
w.r.t. & $\mbf{T}_k \quad (\forall k)$ \\
s.t. & $\mbf{T}_k \in SE(3) \quad (\forall k)$
\end{tabular} \hspace*{-0.2in},
\end{equation}
for some weight matrices, $\mbf{W}_k$ and $\mbf{W}_{k+1,k}$.  Figure~\ref{fig:trajest} depicts the estimation problem as a factor graph.  Similarly to the pose averaging problem, we introduce new optimization variables for the residual errors:
\beqn{}
\mbs{\xi}_k & = & \cay^{-1}\left(\mbf{T}_k\tilde{\mbf{T}}_k^{-1} \right)^\vee, \\
\mbs{\xi}_{k+1,k} & = & \cay^{-1}\left(\mbf{T}_{k+1} \mbf{T}_k^{-1}\tilde{\mbf{T}}_{k+1,k}^{-1} \right)^\vee,
\eeqn
so that the optimization problem can be stated as a \ac{QCQP}:
\begin{equation}\label{eq:trajestprob2}
\begin{tabular}{rl}
$\min$ & $\sum_{k=1}^K \mbs{\xi}_k^T \mbf{W}_k \mbs{\xi}_k + \sum_{k=1}^{K-1} \mbs{\xi}_{k+1,k}^T \mbf{W}_{k+1,k} \mbs{\xi}_{k+1,k}$ \\
w.r.t. & $\mbf{T}_k, \mbs{\xi}_k, \mbs{\xi}_{k+1,k} \quad (\forall k)$ \\
s.t. & $\mbf{C}_k^T \mbf{C}_k = \mbf{I} \quad (\forall k)$ \\
& $\left(\mbf{I} - \frac{1}{2} \mbs{\xi}_k^\wdg \right) \mbf{T}_k = \left(\mbf{I} + \frac{1}{2} \mbs{\xi}_k^\wdg \right) \tilde{\mbf{T}}_k \quad (\forall k)$ \\
& $\left(\mbf{I} - \frac{1}{2} \mbs{\xi}_{k+1,k}^\wdg \right) \mbf{T}_{k+1}$ \\ & $\qquad\qquad = \left(\mbf{I} + \frac{1}{2} \mbs{\xi}_{k+1,k}^\wdg \right) \tilde{\mbf{T}}_{k+1,k} \mbf{T}_k \quad (\forall k)$ 
\end{tabular} \hspace*{-0.2in},
\end{equation}
where the $\det(\mbf{C}_k) = 1$ constraints have been dropped.  Decomposing the matrices according to
\begin{subequations}
\begin{gather}
\mbf{T}_k = \bbm \mbf{C}_k & \mbf{r}_k \\ \mbf{0}^T & 1 \ebm = \bbm \mbf{c}_{k,1} & \mbf{c}_{k,2} & \mbf{c}_{k,3} & \mbf{r}_k \\ 0 & 0 & 0 & 1 \ebm, \\ \tilde{\mbf{T}}_k = \bbm \tilde{\mbf{C}}_k & \tilde{\mbf{r}}_k \\ \mbf{0}^T & 1 \ebm = \bbm \tilde{\mbf{c}}_{k,1} & \tilde{\mbf{c}}_{k,2} & \tilde{\mbf{c}}_{k,2} & \tilde{\mbf{r}}_k \\ 0 & 0 & 0 & 1 \ebm,  \\
\tilde{\mbf{T}}_{k+1,k} = \bbm \tilde{\mbf{C}}_{k+1,k} & \tilde{\mbf{r}}_{k+1,k} \\ \mbf{0}^T & 1 \ebm \qquad\qquad\qquad \\ \qquad = \bbm \tilde{\mbf{c}}_{k+1,k,1} & \tilde{\mbf{c}}_{k+1,k,2} & \tilde{\mbf{c}}_{k+1,k,2} & \tilde{\mbf{r}}_{k+1,k} \\ 0 & 0 & 0 & 1 \ebm, \\ \mbs{\xi}_k = \bbm \mbs{\rho}_k \\ \mbs{\phi}_k \ebm, \quad \mbs{\xi}_{k+1,k} = \bbm \mbs{\rho}_{k+1,k} \\ \mbs{\phi}_{k+1,k} \ebm,
\end{gather}
\end{subequations}
the \ac{QCQP} optimization problem can be rewritten compactly as
\begin{equation}\label{eq:trajestprob3}
\begin{tabular}{rl}
$\min$ & $\sum_{k=1}^K \mbs{\xi}_k^T \mbf{W}_k \mbs{\xi}_k + \sum_{k=1}^{K-1} \mbs{\xi}_{k+1,k}^T \mbf{W}_{k+1,k} \mbs{\xi}_{k+1,k}$  \\
w.r.t. & $\mbf{c}_{k,i}, \mbf{r}_k, \mbs{\rho}_k, \mbs{\phi}_k, \mbs{\rho}_{k+1,k}, \mbs{\phi}_{k+1,k} \quad (\forall i, k)$ \\
s.t. & $\mbf{c}_{k,i}^T\mbf{c}_{k,j} = \delta_{ij} \quad (\forall i,j,k)$ \\
& $(\mbf{I} -  \frac{1}{2}\mbs{\phi}_k^\wdg) \mbf{c}_{k,i} = (\mbf{I} -  \frac{1}{2}\mbs{\phi}_k^\wdg) \tilde{\mbf{c}}_{k,i} \quad (\forall i,k)$ \\
& $(\mbf{I} -  \frac{1}{2}\mbs{\phi}_k^\wdg) \mbf{r}_k = (\mbf{I} +  \frac{1}{2}\mbs{\phi}_k^\wdg) \tilde{\mbf{r}}_k + \mbs{\rho}_k \quad (\forall k)$ \\
& $(\mbf{I} -  \frac{1}{2}\mbs{\phi}_{k+1,k}^\wdg) \mbf{c}_{k+1,i} $ \\ & $ \qquad = (\mbf{I} -  \frac{1}{2}\mbs{\phi}_{k+1,k}^\wdg) \tilde{\mbf{C}}_{k+1,k} \mbf{c}_{k,i} \quad (\forall i,k)$ \\
& $(\mbf{I} -  \frac{1}{2}\mbs{\phi}_{k+1,k}^\wdg) \mbf{r}_{k+1} $ \\ & $ \qquad = (\mbf{I} +  \frac{1}{2}\mbs{\phi}_{k+1,k}^\wdg)\left( \tilde{\mbf{C}}_{k+1,k} \mbf{r}_k + \tilde{\mbf{r}}_{k+1,k} \right) $ \\ & $ \qquad\qquad  + \; \mbs{\rho}_{k+1,k} \quad (\forall k)$
\end{tabular}.
\end{equation}
Similarly to the pose averaging problem, if we convert this \ac{QCQP} to a \ac{SDP}, it is not \change{always tight even for low noise levels. }  We need to include some redundant constraints to \change{improve tightness. }  For each of the $\mbs{\xi}_k$ variables, we can create copies of the redundant constraints required in the pose averaging problem.  However, this is still not enough; we require some additional constraints involving the $\mbs{\xi}_{k+1,k}$ variables.  

\begin{figure*}[t]
\centering
\includegraphics[width=\textwidth]{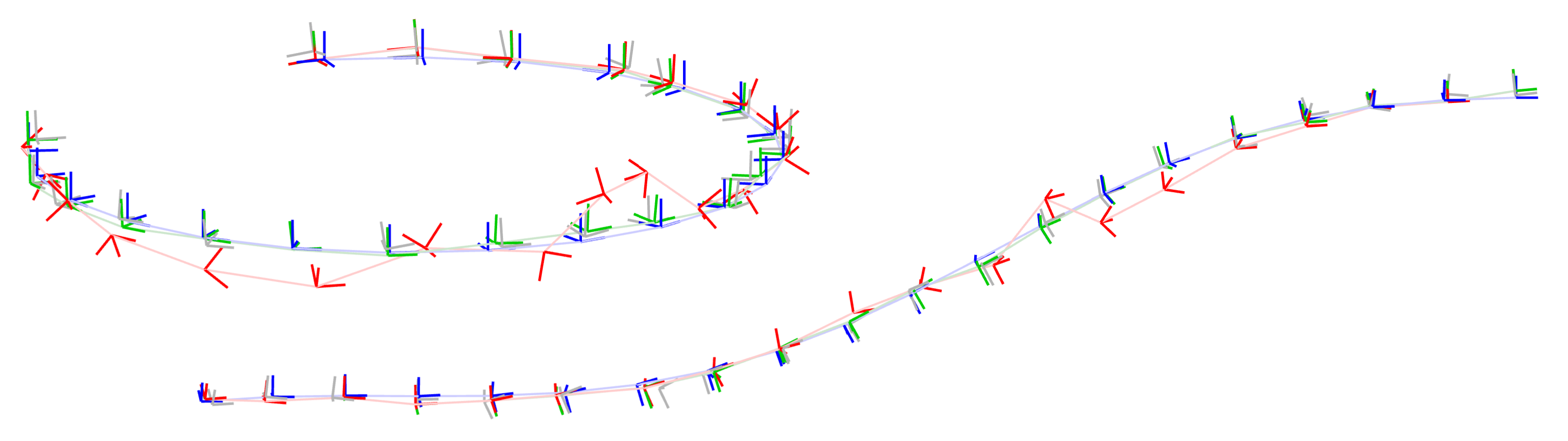}
\caption{{\em Discrete-time Trajectory Estimation:}  Two examples of discrete-time trajectory estimation where the randomly initialized local solver (red) becomes trapped in a poor local minimum while the global solver (green) finds the correct solution, which is closer to the groundtruth (blue).  The noisy pose measurements are also shown (grey).  It is interesting to note that the poor local solver solutions are twisted around the groundtruth.}
\label{fig:cayley_trajest3}
\end{figure*}

Such additional redundant constraints can be formed by combining the last two of~\eqref{eq:trajestprob3}; the second last can be written as
\begin{multline}
\frac{1}{2}  ( \mbf{c}_{k+1,i} + \tilde{\mbf{C}}_{k+1,k} \mbf{c}_{k,i} )^T \mbs{\phi}_{k+1,k}^\wdg \\ = -\left(\mbf{c}_{k+1,i} - \tilde{\mbf{C}}_{k+1,k} \mbf{c}_{k,i}\right)^T ,
\end{multline}
while the last can be premultiplied by $( \mbf{c}_{k+1,i} + \tilde{\mbf{C}}_{k+1,k} \mbf{c}_{k,i} )^T$ and written as
\begin{multline}
( \mbf{c}_{k+1,i} + \tilde{\mbf{C}}_{k+1,k} \mbf{c}_{k,i} )^T ( \mbf{r}_{k+1} - \tilde{\mbf{C}}_{k+1,k} \mbf{r}_k - \tilde{\mbf{r}}_{k+1,k} ) \\ = \underbrace{\frac{1}{2} ( \mbf{c}_{k+1,i} + \tilde{\mbf{C}}_{k+1,k} \mbf{c}_{k,i} )^T \mbs{\phi}_{k+1,k}^\wdg}_{-(\mbf{c}_{k+1,i} -  \tilde{\mbf{C}}_{k+1,k} \mbf{c}_{k,i})^T} \\ \times \; ( \mbf{r}_{k+1} + \tilde{\mbf{C}}_{k+1,k} \mbf{r}_k +\tilde{\mbf{r}}_{k+1,k} ) \\ + ( \mbf{c}_{k+1,i} + \tilde{\mbf{C}}_{k+1,k} \mbf{c}_{k,i} )^T \mbs{\rho}_{k+1,k}.
\end{multline}
After performing the indicated substitution, this becomes
\begin{multline}
\frac{1}{2} (\mbf{c}_{k+1,i} + \tilde{\mbf{C}}_{k+1,k} \mbf{c}_{k,i} )^T \mbs{\rho}_{k+1,k} \\ = \mbf{c}_{k+1,i}^T \mbf{r}_{k+1} - \mbf{c}_{k,i}^T ( \mbf{r}_{k} + \tilde{\mbf{C}}_{k+1,k}^T \tilde{\mbf{r}}_{k+1,k}),
\end{multline}
which is once again a quadratic constraint.

\begin{figure*}[t]
\centering
\includegraphics[width=0.47\textwidth]{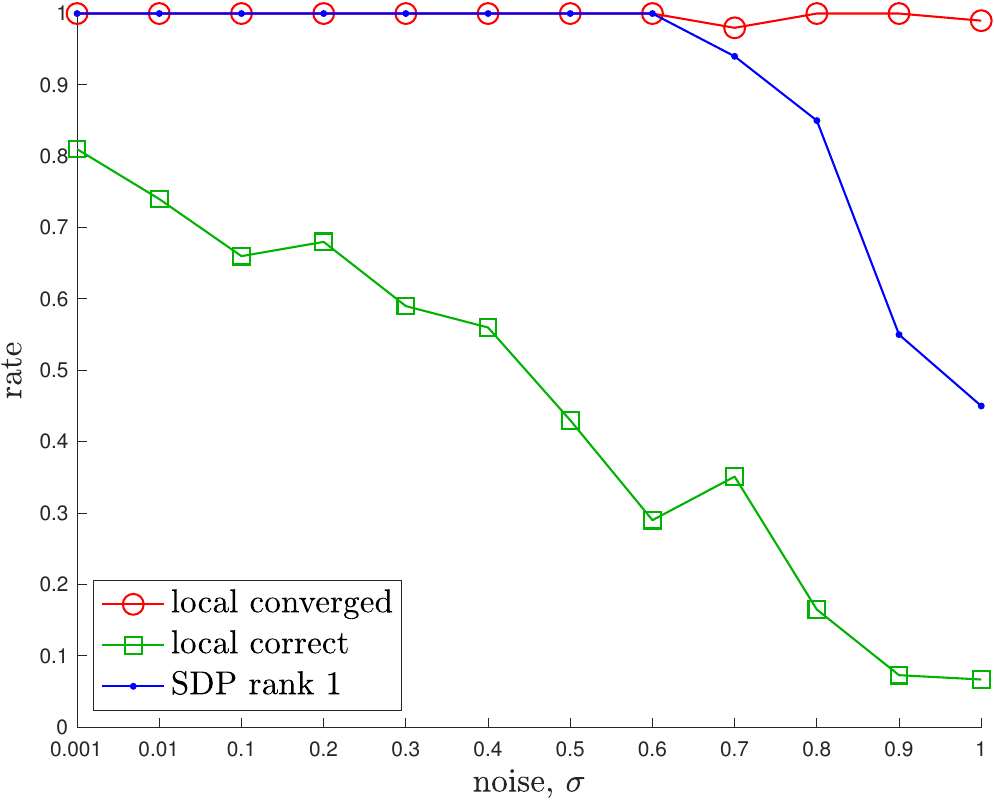}\hspace{0.2in}
\includegraphics[width=0.47\textwidth]{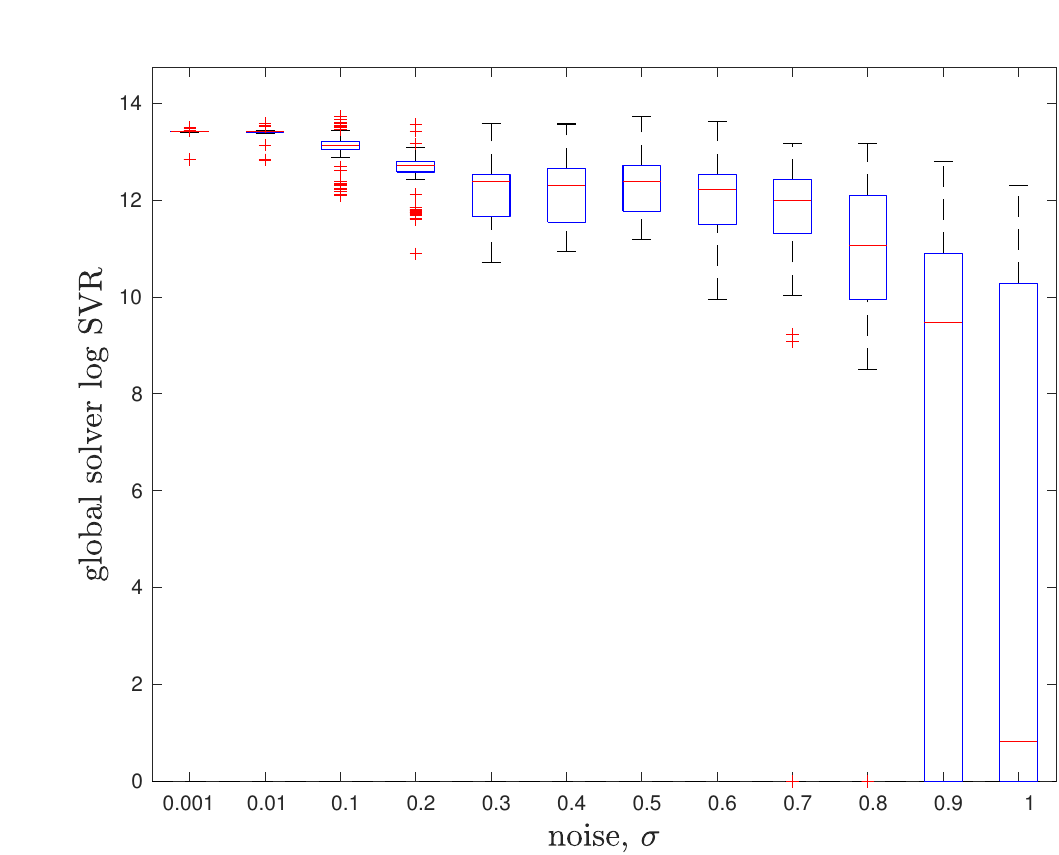} 
\caption{{\em Discrete-time Trajectory Estimation:} A quantitative evaluation of the tightness of the discrete-time trajectory estimation problem with increasing measurement noise, $\sigma$.  At each noise level, we conducted $100$ trials with the geometry of the trajectory as in the left example of Figure~\ref{fig:cayley_trajest3}.  \change{(left) We see that the local solver (randomly initialized) finds the global minimum with decreasing frequency (green) as the measurement noise is increased, while the \ac{SDP} solver (blue) successfully produces rank-1 solutions (we consider log \ac{SVR} of at least $5$ to be rank 1) to much higher noise levels.  For completeness, we also show how frequently the local solver converges to any minimum (red). } (right) Boxplots of the log \ac{SVR} of the \ac{SDP} solution show that the global solution remains highly rank 1 over a wide range of measurement noise values.}
\label{fig:cayley_trajest3_noise}
\end{figure*}

Summarizing, the following \ac{QCQP} offers a reasonably tight \ac{SDP} relaxation in practice:
\begin{equation}\label{eq:trajestprob4}
\begin{tabular}{rl}
$\min$ & $\sum_{k=1}^K \mbs{\xi}_k^T \mbf{W}_k \mbs{\xi}_k + \sum_{k=1}^{K-1} \mbs{\xi}_{k+1,k}^T \mbf{W}_{k+1,k} \mbs{\xi}_{k+1,k}$  \\
w.r.t. & $\mbf{c}_{k,i}, \mbf{r}_k, \mbs{\rho}_k, \mbs{\phi}_k, \mbs{\rho}_{k+1,k}, \mbs{\phi}_{k+1,k} \quad (\forall i, k)$ \\
s.t. & $\mbf{c}_{k,i}^T\mbf{c}_{k,j} = \delta_{ij} \quad (\forall i,j,k)$ \\
& $(\mbf{I} -  \frac{1}{2}\mbs{\phi}_k^\wdg) \mbf{c}_{k,i} = (\mbf{I} -  \frac{1}{2}\mbs{\phi}_k^\wdg) \tilde{\mbf{c}}_{k,i} \quad (\forall i,k)$ \\
& $(\mbf{I} -  \frac{1}{2}\mbs{\phi}_k^\wdg) \mbf{r}_k = (\mbf{I} +  \frac{1}{2}\mbs{\phi}_k^\wdg) \tilde{\mbf{r}}_k + \mbs{\rho}_k \quad (\forall k)$ \\
& $(\mbf{I} -  \frac{1}{2}\mbs{\phi}_{k+1,k}^\wdg) \mbf{c}_{k+1,i} $ \\ & $ \qquad = (\mbf{I} -  \frac{1}{2}\mbs{\phi}_{k+1,k}^\wdg) \tilde{\mbf{C}}_{k+1,k} \mbf{c}_{k,i} \quad (\forall i,k)$ \\
& $(\mbf{I} -  \frac{1}{2}\mbs{\phi}_{k+1,k}^\wdg) \mbf{r}_{k+1} $ \\ & $\qquad = (\mbf{I} +  \frac{1}{2}\mbs{\phi}_{k+1,k}^\wdg)\left( \tilde{\mbf{C}}_{k+1,k} \mbf{r}_k + \tilde{\mbf{r}}_{k+1,k} \right) $ \\ & $ \qquad\quad  + \; \mbs{\rho}_{k+1,k} \quad (\forall k)$ \\
(red.) & $ \frac{1}{2} ( \mbf{c}_{k,i} + \tilde{\mbf{c}}_{k,i} )^T \mbs{\rho}_k = \mbf{c}_{k,i}^T \mbf{r}_k - \tilde{\mbf{c}}_{k,i}^T \tilde{\mbf{r}}_k \quad (\forall i,k)$ \\
& $\mbf{r}_k^T\mbf{r}_k  = \mbf{r}_k^T \tilde{\mbf{r}}_k  -  \frac{1}{2} \mbf{r}_k^T  \tilde{\mbf{r}}_k^\wdg \mbs{\phi}_k + \mbf{r}_k^T\mbs{\rho}_k \quad (\forall k)$ \\
& $\frac{1}{2} (\mbf{c}_{k+1,i} + \tilde{\mbf{C}}_{k+1,k} \mbf{c}_{k,i} )^T \mbs{\rho}_{k+1,k} $ \\ & $\qquad = \mbf{c}_{k+1,i}^T \mbf{r}_{k+1} $ \\ & $ \qquad\quad - \; \mbf{c}_{k,i}^T ( \mbf{r}_{k} + \tilde{\mbf{C}}_{k+1,k}^T \tilde{\mbf{r}}_{k+1,k}) \quad (\forall i,k)$
\end{tabular}\hspace*{-0.2in}.
\end{equation}
We again leave it to the reader to manipulate this into the standard form of~\eqref{eq:sdp2}.

\begin{figure*}[t]
\centering
\includegraphics[width=0.59\textwidth]{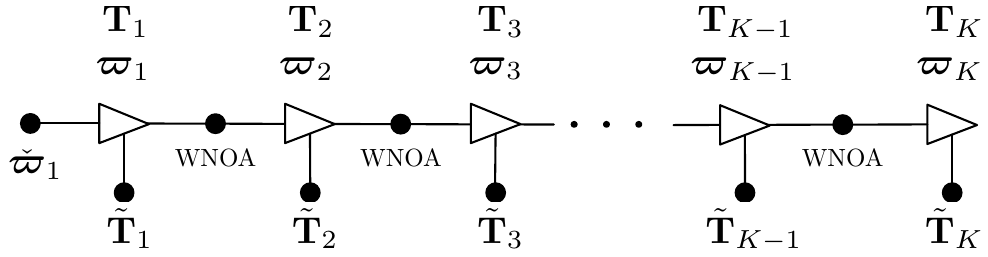}
\caption{Factor graph representation of the continuous-time estimation problem.  Each block dot represents one of the error terms in the cost function of~\eqref{eq:cttrajestprob}.}
\label{fig:cttrajest}
\end{figure*}

The appendix provides a baseline local solver for this problem.  For the global (SDP) solver we used \texttt{cvx} in Matlab with \texttt{mosek} \citep{mosek}.  The solution costs of the global and local solvers agree to high precision if a good initial guess is given to the local solver.  Figure~\ref{fig:cayley_trajest3} provides examples of the local solver becoming trapped in poor local minima while the global solver converges to the correct minima near the groundtruth.  Figure~\ref{fig:cayley_trajest3_noise} provides a quantitative study of the tightness of the \ac{SDP} solution with increasing measurement noise; we selected the measurement covariances as $\mbf{W}_k^{-1} = \mbf{W}_{k+1,k}^{-1} = \sigma^2 \mbf{I}$, with $\sigma$ increasing.  We again see there is a large range for the noise over which the local solver can become trapped in a local minimum while the global solver remains tight; in fact, even at very low noise levels it is quite easy to have the local solver become trapped.  With $K = 20$ poses in the trajectory, the local solver took on average \change{$0.1574s$ while the SDP solver took on average $14.32s$.}


\section{Continuous-Time Trajectory Estimation}
\label{sec:continuoustime}

Finally, we consider so-called continuous-time trajectory estimation.  Continuous-time methods come in parametric \citep{furgale_icra12} and nonparametric \citep{barfoot_rss14, anderson_iros15} varieties; here we will discuss the latter.  We consider a continuous-time \ac{GP} prior over the trajectory known as \ac{WNOA}; this serves to smooth the trajectory and is fused with pose measurements provided at discrete times.  We will still ultimately have to discretize the trajectory for the purpose of estimation and so will have $K$ states comprising both pose and generalized velocity (a.k.a., twist), $\left\{ \mbf{T}_k, \mbs{\varpi}_k \right\}$.  Figure~\ref{fig:cttrajest} depicts the situation as a factor graph.  In practice, we may not actually have pose measurements at every time at which we introduce a state variable.

\begin{figure*}[t]
\centering
\includegraphics[width=\textwidth]{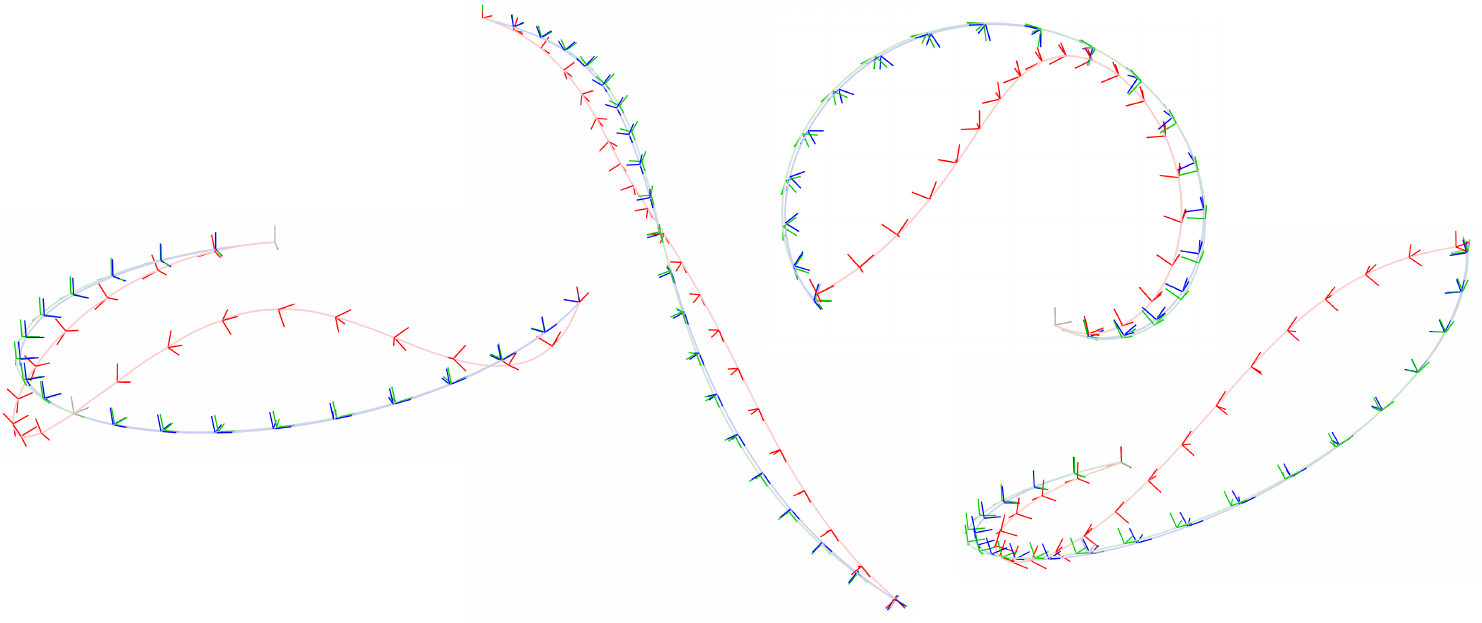}
\caption{{\em Continuous-time Trajectory Estimation:} Four examples of continuous-time trajectory estimation where the randomly initialized local solver (red) becomes trapped in a poor local minimum while the global solver (green) finds the correct solution, which is closer to the groundtruth (blue).  The noisy pose measurements (occurring only at the start, middle, and end of each trajectory) are also shown (grey).  The local minimum in the leftmost example is very similar to one reported by \citet{lilge_ijrr22}.}
\label{fig:cayley_continuum}
\end{figure*}

The optimization problem that we want to solve in this case is
\begin{equation}\label{eq:cttrajestprob}
\begin{tabular}{rl}
$\min$ & $\sum_{k=1}^K \cay^{-1}\left(\mbf{T}_k\tilde{\mbf{T}}_k^{-1} \right)^{\vee^T} \! \mbf{W}_k \, \cay^{-1}\left(\mbf{T}_k\tilde{\mbf{T}}_k^{-1} \right)^\vee$ \\ & $\qquad + \; (\pri{\mbs{\varpi}}_1 - \mbs{\varpi}_1)^T \mbf{Q}^{-1}_1 (\pri{\mbs{\varpi}}_1 - \mbs{\varpi}_1)$ \\ & $\qquad \qquad + \; \sum_{k=1}^{K-1} \mbf{e}_{k+1,k}^T \! \mbf{Q}^{-1}_{k+1,k} \, \mbf{e}_{k+1,k}$ \\
w.r.t. & $\mbf{T}_k, \mbs{\varpi}_k \quad (\forall k)$ \\
s.t. & $\mbf{T}_k \in SE(3) \quad (\forall k)$
\end{tabular}\hspace*{-0.15in},
\end{equation}
for some weight matrices, $\mbf{W}_k$, and 
\begin{subequations}
\begin{gather}\label{eq:gpprior}
\mbf{e}_{k+1,k} = \bbm (t_{k+1} - t_k) \mbs{\varpi}_k - \cay^{-1}\left(\mbf{T}_{k+1} \mbf{T}_k^{-1} \right)^\vee \\ \mbs{\varpi}_k - \mbs{\varpi}_{k+1} \ebm, \\ \mbf{Q}_{k+1,k} = \bbm \frac{1}{3}(t_{k+1}-t_k)^3 \mbf{Q}_c & \frac{1}{2}(t_{k+1}-t_k)^2 \mbf{Q}_c  \\ \frac{1}{2}(t_{k+1}-t_k)^2 \mbf{Q}_c  & (t_{k+1} -t_k) \mbf{Q}_c \ebm.
\end{gather}
\end{subequations}
The $t_k$ are known timestamps of the states , $\mbf{Q}_c$ is a power-spectral density matrix affecting smoothness of the \ac{GP} prior, and $\pri{\mbs{\varpi}}_1$ together with $\mbf{Q}_1$ represent a Gaussian prior on the initial generalized velocity.  The \ac{GP} prior defined by~\eqref{eq:gpprior} is similar in spirit to the one first defined by \citet{anderson_iros15}, only now adapted to work with the Cayley map.  Looking at $\mbf{e}_{k+1,k}$, the first row encourages the change in pose from one time to the next to be similar to the generalized velocity multiplied by the change in time; the second row encourages the generalized velocity to remain constant over time (i.e., no acceleration).   The process noise covariance, $\mbf{Q}_{k+1,k}$, comes from integrating the \ac{WNOA} prior over the same time interval \citep{barfoot_rss14, barfoot_ser24}.

Similarly to the discrete-time trajectory estimation case, we introduce the following substitution variables\footnote{Importantly, we were forced to deviate from our program of substitution variables being the residual errors of the cost function in our choice of $\mbs{\xi}_{k+1,k}$; this could be problematic for very-high-velocity trajectories as $\mbs{\xi}_{k+1,k}$ could become large and approach the singularity of the Cayley map.  We justify our choice as providing the necessary pathway to a \ac{QCQP}, but this could be revisited in future work.}:
\beqn{}
\mbs{\xi}_k & = & \cay^{-1}\left(\mbf{T}_k\tilde{\mbf{T}}_k^{-1} \right)^\vee, \\
\mbs{\xi}_{k+1,k} & = & \cay^{-1}\left(\mbf{T}_{k+1} \mbf{T}_k^{-1} \right)^\vee,
\eeqn
so that the optimization problem can be stated as a \ac{QCQP}:
\begin{equation}\label{eq:cttrajestprob2}
\begin{tabular}{rl}
$\min$ & $\sum_{k=1}^K \mbs{\xi}_k^T \mbf{W}_k \mbs{\xi}_k + (\pri{\mbs{\varpi}}_1 - \mbs{\varpi}_1)^T \mbf{Q}^{-1}_1 (\pri{\mbs{\varpi}}_1 - \mbs{\varpi}_1)$ \\ & $\qquad + \; \sum_{k=1}^{K-1} \bbm (t_{k+1} - t_k) \mbs{\varpi}_k - \mbs{\xi}_{k+1,k} \\ \mbs{\varpi}_k - \mbs{\varpi}_{k+1} \ebm^T $ \\ & $ \qquad\quad \times \; \mbf{Q}^{-1}_{k+1,k} \, \bbm (t_{k+1} - t_k) \mbs{\varpi}_k - \mbs{\xi}_{k+1,k} \\ \mbs{\varpi}_k - \mbs{\varpi}_{k+1} \ebm$ \\
w.r.t. & $\mbf{T}_k, \mbs{\varpi}_k, \mbs{\xi}_k, \mbs{\xi}_{k+1,k} \quad (\forall k)$ \\
s.t. & $\mbf{C}_k^T \mbf{C}_k = \mbf{I} \quad (\forall k)$ \\
& $\left(\mbf{I} - \frac{1}{2} \mbs{\xi}_k^\wdg \right) \mbf{T}_k = \left(\mbf{I} + \frac{1}{2} \mbs{\xi}_k^\wdg \right) \tilde{\mbf{T}}_k \quad (\forall k)$ \\
& $\left(\mbf{I} - \frac{1}{2} \mbs{\xi}_{k+1,k}^\wdg \right) \mbf{T}_{k+1} = \left(\mbf{I} + \frac{1}{2} \mbs{\xi}_{k+1,k}^\wdg \right) \mbf{T}_k \quad (\forall k)$ 
\end{tabular}\hspace*{-0.2in},
\end{equation}
where the $\det(\mbf{C}_k) = 1$ constraints have been dropped.   Decomposing the matrices according to
\begin{subequations}
\begin{gather}
\mbf{T}_k = \bbm \mbf{C}_k & \mbf{r}_k \\ \mbf{0}^T & 1 \ebm = \bbm \mbf{c}_{k,1} & \mbf{c}_{k,2} & \mbf{c}_{k,3} & \mbf{r}_k \\ 0 & 0 & 0 & 1 \ebm, \\ \tilde{\mbf{T}}_k = \bbm \tilde{\mbf{C}}_k & \tilde{\mbf{r}}_k \\ \mbf{0}^T & 1 \ebm = \bbm \tilde{\mbf{c}}_{k,1} & \tilde{\mbf{c}}_{k,2} & \tilde{\mbf{c}}_{k,2} & \tilde{\mbf{r}}_k \\ 0 & 0 & 0 & 1 \ebm,  \\ \mbs{\xi}_k = \bbm \mbs{\rho}_k \\ \mbs{\phi}_k \ebm, \quad \mbs{\xi}_{k+1,k} = \bbm \mbs{\rho}_{k+1,k} \\ \mbs{\phi}_{k+1,k} \ebm,
\end{gather}
\end{subequations}
the \ac{QCQP} optimization problem can be rewritten compactly as
\begin{equation}\label{eq:cttrajestprob3}
\begin{tabular}{rl}
$\min$ & $\sum_{k=1}^K \mbs{\xi}_k^T \mbf{W}_k \mbs{\xi}_k + (\pri{\mbs{\varpi}}_1 - \mbs{\varpi}_1)^T \mbf{Q}^{-1}_1 (\pri{\mbs{\varpi}}_1 - \mbs{\varpi}_1)$ \\ & $\qquad + \; \sum_{k=1}^{K-1} \bbm (t_{k+1} - t_k) \mbs{\varpi}_k - \mbs{\xi}_{k+1,k} \\ \mbs{\varpi}_k - \mbs{\varpi}_{k+1} \ebm^T $ \\ & $ \qquad\quad \times \; \mbf{Q}^{-1}_{k+1,k} \, \bbm (t_{k+1} - t_k) \mbs{\varpi}_k - \mbs{\xi}_{k+1,k} \\ \mbs{\varpi}_k - \mbs{\varpi}_{k+1} \ebm$ \\
w.r.t. & $\mbf{c}_{k,i}, \mbf{r}_k, \mbs{\rho}_k, \mbs{\phi}_k, \mbs{\rho}_{k+1,k}, \mbs{\phi}_{k+1,k}, \mbs{\varpi}_k \quad (\forall i, k)$ \\
s.t. & $\mbf{c}_{k,i}^T\mbf{c}_{k,j} = \delta_{ij} \quad (\forall i,j,k)$ \\
& $(\mbf{I} -  \frac{1}{2}\mbs{\phi}_k^\wdg) \mbf{c}_{k,i} = (\mbf{I} -  \frac{1}{2}\mbs{\phi}_k^\wdg) \tilde{\mbf{c}}_{k,i} \quad (\forall i,k)$ \\
& $(\mbf{I} -  \frac{1}{2}\mbs{\phi}_k^\wdg) \mbf{r}_k = (\mbf{I} +  \frac{1}{2}\mbs{\phi}_k^\wdg) \tilde{\mbf{r}}_k + \mbs{\rho}_k \quad (\forall k)$ \\
& $(\mbf{I} -  \frac{1}{2}\mbs{\phi}_{k+1,k}^\wdg) \mbf{c}_{k+1,i} $ \\ & $ \qquad = \; (\mbf{I} -  \frac{1}{2}\mbs{\phi}_{k+1,k}^\wdg)\mbf{c}_{k,i} \quad (\forall i,k)$ \\
& $(\mbf{I} -  \frac{1}{2}\mbs{\phi}_{k+1,k}^\wdg) \mbf{r}_{k+1} $ \\ & $ \qquad = \; (\mbf{I} +  \frac{1}{2}\mbs{\phi}_{k+1,k}^\wdg)\mbf{r}_k  + \mbs{\rho}_{k+1,k} \quad (\forall k)$
\end{tabular}\hspace*{-0.2in}.
\end{equation}
Similarly to the problems above, if we convert this \ac{QCQP} to a \ac{SDP}, it is not tight \change{even for low noise levels. }  We need to include some redundant constraints to \change{improve tightness. }  For each of the $\mbs{\xi}_k$  and $\mbs{\xi}_{k+1,k}$ variables, we can create copies of the redundant constraints required in the discrete-time trajectory estimation problem.  However, this is still not always enough to tighten the problem, particularly when we do not have a pose measurement at every state time (see below comment regarding sparser measurement graphs).  

We can generate some additional redundant constraints fairly easily.  First, we can premultiply the second constraint of~\eqref{eq:cttrajestprob3} by $\mbs{\phi}_k^T$ so that
\begin{multline}
\mbs{\phi}_k^T(\mbf{I} -  \frac{1}{2}\mbs{\phi}_k^\wdg) \mbf{c}_{k,i} = \mbs{\phi}_k^T(\mbf{I} -  \frac{1}{2}\mbs{\phi}_k^\wdg) \tilde{\mbf{c}}_{k,i} \\ \Rightarrow \quad \mbs{\phi}_k^T\mbf{c}_{k,i}  = \mbs{\phi}_k^T\tilde{\mbf{c}}_{k,i},
\end{multline}
where we have used that $\mbs{\phi}_k^T \mbs{\phi}_k^\wdg = \mbf{0}^T$.
Similarly, premultiplying the third constraint in~\eqref{eq:cttrajestprob3} by $\mbs{\phi}_k^T$ results in
\begin{equation}
\mbs{\phi}_k^T \mbf{r}_k = \mbs{\phi}_k^T \tilde{\mbf{r}}_k + \mbs{\phi}_k^T \mbs{\rho}_k.
\end{equation}
Premultiplying the fourth and fifth constraints in~\eqref{eq:cttrajestprob3} by $\mbs{\phi}_{k+1,k}^T$ results in
\begin{subequations}
\begin{gather}
\mbs{\phi}_{k+1,k}^T\mbf{c}_{k+1,i}  = \mbs{\phi}_{k+1,k}^T\mbf{c}_{k,i}, \\ \mbs{\phi}_{k+1,k}^T \mbf{r}_{k+1} = \mbs{\phi}_{k+1,k}^T \mbf{r}_k + \mbs{\phi}_{k+1,k}^T \mbs{\rho}_{k+1,k}.
\end{gather}
\end{subequations}
Next, we can exploit the fact that columns of a rotation matrix satisfy $\mbf{c}_{\ell}^\wdg \mbf{c}_{m} = \mbf{c}_n$ where $\ell m n \in \left\{123, 231, 312 \right\}$.  If we premultiply the second constraint of~\eqref{eq:cttrajestprob3} by $\mbf{c}_{k,m}^T$ we have
\begin{multline}
\mbf{c}_{k,m}^T (\mbf{I} -  \frac{1}{2}\mbs{\phi}_k^\wdg) \mbf{c}_{k,\ell} = \mbf{c}_{k,m}^T(\mbf{I} -  \frac{1}{2}\mbs{\phi}_k^\wdg) \tilde{\mbf{c}}_{k,\ell} \\ \Rightarrow \quad \mbf{c}_{k,m}^T\mbf{c}_{k,\ell} - \frac{1}{2} \mbs{\phi}_k^T \mbf{c}_{k,n}  = \mbf{c}_{k,m}^T\tilde{\mbf{c}}_{k,\ell} - \frac{1}{2} \mbf{c}_{k,m}^T \tilde{\mbf{c}}_{k,\ell}^\wdg \mbs{\phi}_k,
\end{multline}
which is still a quadratic constraint.  Finally, if we premultiply the last constraint of~\eqref{eq:cttrajestprob3} by $(\mbf{r}_{k+1} + \mbf{r}_k)^T$, this results in 
\begin{equation}
\mbf{r}_{k+1}^T \mbf{r}_{k+1} = \mbf{r}_{k+1}^T \mbs{\rho}_{k+1,k} + \mbf{r}_k^T \mbs{\rho}_{k+1,k}  + \mbf{r}_k^T \mbf{r}_k,
\end{equation}
which is once again a quadratic constraint.

\begin{figure*}[t]
\centering
\includegraphics[height=0.38\textwidth]{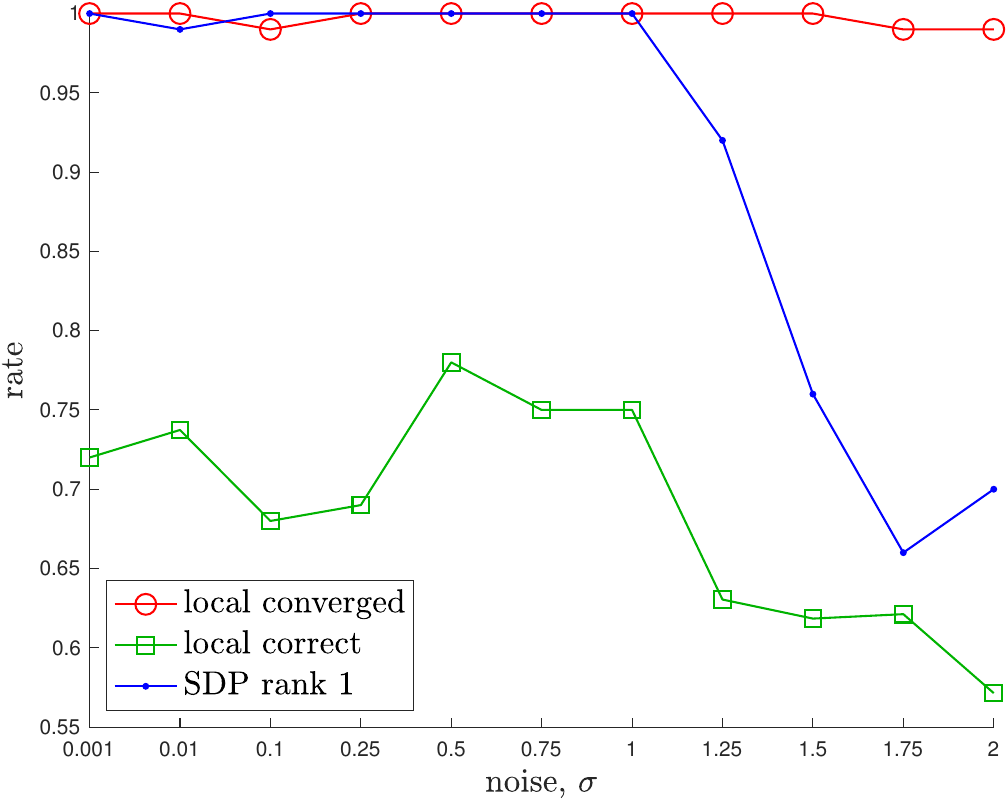}\hspace{0.2in}
\includegraphics[height=0.377\textwidth]{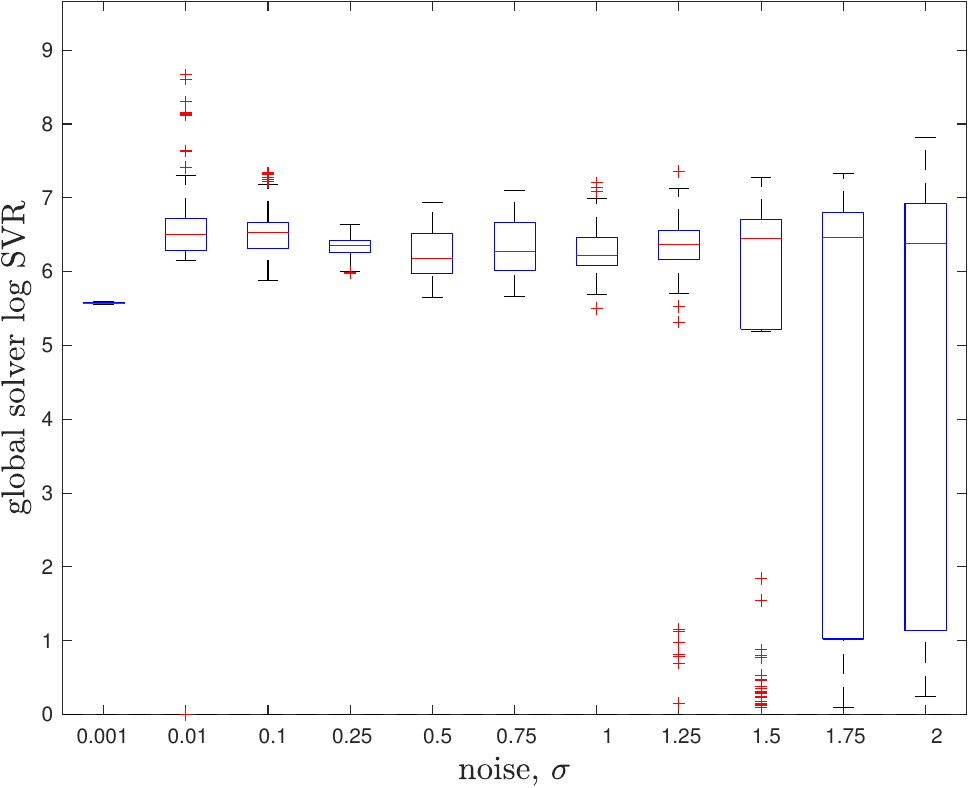} 
\caption{{\em Continuous-time Trajectory Estimation:} A quantitative evaluation of the tightness of the continuous-time trajectory estimation problem with increasing measurement noise, $\sigma$.  At each noise level, we conducted $100$ trials with the geometry of the trajectory as in the left example of Figure~\ref{fig:cayley_continuum}.  \change{(left) We see that the local solver (randomly initialized) finds the global minimum with decreasing frequency (green) as the measurement noise is increased, while the \ac{SDP} solver (blue) successfully produces rank-1 solutions (we consider log \ac{SVR} of at least $5$ to be rank 1) to much higher noise levels.  For completeness, we also show how frequently the local solver converges to any minimum (red). } (right) Boxplots of the log \ac{SVR} of the \ac{SDP} solution show that the global solution remains reasonably rank 1 over a wide range of measurement noise values.}
\label{fig:cayley_continuum_noise}
\end{figure*}

Summarizing, the following \ac{QCQP} offers a reasonably tight \ac{SDP} relaxation in practice:
\begin{equation}\label{eq:trajestprob4}
\begin{tabular}{rl}
$\min$ & $\sum_{k=1}^K \mbs{\xi}_k^T \mbf{W}_k \mbs{\xi}_k + (\pri{\mbs{\varpi}}_1 - \mbs{\varpi}_1)^T \mbf{Q}^{-1}_1 (\pri{\mbs{\varpi}}_1 - \mbs{\varpi}_1)$ \\ & $\qquad + \; \sum_{k=1}^{K-1} \bbm (t_{k+1} - t_k) \mbs{\varpi}_k - \mbs{\xi}_{k+1,k} \\ \mbs{\varpi}_k - \mbs{\varpi}_{k+1} \ebm^T $ \\ & $ \qquad\quad \times \; \mbf{Q}^{-1}_{k+1,k} \, \bbm (t_{k+1} - t_k) \mbs{\varpi}_k - \mbs{\xi}_{k+1,k} \\ \mbs{\varpi}_k - \mbs{\varpi}_{k+1} \ebm$ \\
w.r.t. & $\mbf{c}_{k,i}, \mbf{r}_k, \mbs{\rho}_k, \mbs{\phi}_k, \mbs{\rho}_{k+1,k}, \mbs{\phi}_{k+1,k}, \mbs{\varpi}_k \quad (\forall i, k)$ \\
s.t. & $\mbf{c}_{k,i}^T\mbf{c}_{k,j} = \delta_{ij} \quad (\forall i,j,k)$ \\
& $(\mbf{I} -  \frac{1}{2}\mbs{\phi}_k^\wdg) \mbf{c}_{k,i} = (\mbf{I} -  \frac{1}{2}\mbs{\phi}_k^\wdg) \tilde{\mbf{c}}_{k,i} \quad (\forall i,k)$ \\
& $(\mbf{I} -  \frac{1}{2}\mbs{\phi}_k^\wdg) \mbf{r}_k = (\mbf{I} +  \frac{1}{2}\mbs{\phi}_k^\wdg) \tilde{\mbf{r}}_k + \mbs{\rho}_k \quad (\forall k)$ \\
& $(\mbf{I} -  \frac{1}{2}\mbs{\phi}_{k+1,k}^\wdg) \mbf{c}_{k+1,i} $ \\ & $ \qquad = \; (\mbf{I} -  \frac{1}{2}\mbs{\phi}_{k+1,k}^\wdg)\mbf{c}_{k,i} \quad (\forall i,k)$ \\
& $(\mbf{I} -  \frac{1}{2}\mbs{\phi}_{k+1,k}^\wdg) \mbf{r}_{k+1} $ \\ & $ \qquad = \; (\mbf{I} +  \frac{1}{2}\mbs{\phi}_{k+1,k}^\wdg)\mbf{r}_k  + \mbs{\rho}_{k+1,k} \quad (\forall k)$
 \\
(red.) & $ \frac{1}{2} ( \mbf{c}_{k,i} + \tilde{\mbf{c}}_{k,i} )^T \mbs{\rho}_k = \mbf{c}_{k,i}^T \mbf{r}_k - \tilde{\mbf{c}}_{k,i}^T \tilde{\mbf{r}}_k \quad (\forall i,k)$ \\
& $\mbf{r}_k^T\mbf{r}_k  = \mbf{r}_k^T \tilde{\mbf{r}}_k  -  \frac{1}{2} \mbf{r}_k^T  \tilde{\mbf{r}}_k^\wdg \mbs{\phi}_k + \mbf{r}_k^T\mbs{\rho}_k \quad (\forall k)$ \\
& $\mbs{\phi}_k^T\mbf{c}_{k,i}  = \mbs{\phi}_k^T\tilde{\mbf{c}}_{k,i} \quad (\forall i,k)$ \\
& $\mbs{\phi}_k^T \mbf{r}_k = \mbs{\phi}_k^T \tilde{\mbf{r}}_k + \mbs{\phi}_k^T \mbs{\rho}_k  \quad (\forall k)$ \\
& $\mbf{c}_{k,m}^T\mbf{c}_{k,\ell} - \frac{1}{2} \mbs{\phi}_k^T \mbf{c}_{k,n}  $ \\ & $ \qquad = \mbf{c}_{k,m}^T\tilde{\mbf{c}}_{k,\ell} - \frac{1}{2} \mbf{c}_{k,m}^T \tilde{\mbf{c}}_{k,\ell}^\wdg \mbs{\phi}_k $ \\ & $ \qquad\qquad\qquad\quad (\forall \ell m n \in \left\{123, 231, 312 \right\}, k)$ \\
& $\frac{1}{2} (\mbf{c}_{k+1,i} + \mbf{c}_{k,i} )^T \mbs{\rho}_{k+1,k} $ \\ & $ \qquad = \; \mbf{c}_{k+1,i}^T \mbf{r}_{k+1} - \mbf{c}_{k,i}^T \mbf{r}_{k} \quad (\forall i,k)$ \\
& $\mbf{r}_{k+1}^T \mbf{r}_{k+1} $ \\ & $ \qquad = \; \mbf{r}_{k+1}^T \mbs{\rho}_{k+1,k} + \mbf{r}_k^T \mbs{\rho}_{k+1,k}  + \mbf{r}_k^T \mbf{r}_k \quad (\forall k)$ \\
& $\mbs{\phi}_{k+1,k}^T\mbf{c}_{k+1,i}  = \mbs{\phi}_{k+1,k}^T\mbf{c}_{k,i} \quad (\forall i,k)$ \\
& $\mbs{\phi}_{k+1,k}^T \mbf{r}_{k+1} = \mbs{\phi}_{k+1,k}^T \mbf{r}_k + \mbs{\phi}_{k+1,k}^T \mbs{\rho}_{k+1,k}  \quad (\forall k)$
\end{tabular}.
\end{equation}
We again leave it to the reader to manipulate this into the standard form of~\eqref{eq:sdp2}.  We can also notice that the $\mbs{\varpi}_k$ variables are not involved in any of the constraints and thus remain unconstrained variables.  Similar to \citet{rosen19,holmes23}, at implementation we use the Schur complement to marginalize these variables out of the problem, thereby keeping the size of the \ac{SDP} as small as possible; we can easily compute them after the main solve.

The appendix provides a baseline local solver for this problem.  For the global (SDP) solver we used \texttt{cvx} in Matlab with \texttt{mosek} \citep{mosek}.  The solution costs of the global and local solvers agree to high precision if a good initial guess is given to the local solver.  Figure~\ref{fig:cayley_continuum} provides examples of the local solver becoming trapped in poor local minima while the global solver converges to the correct minimum near the groundtruth.  Figure~\ref{fig:cayley_continuum_noise} provides a quantitative study of the tightness of the \ac{SDP} solution with increasing measurement noise; we selected the measurement covariances as $\mbf{W}_k^{-1} = \sigma^2 \mbf{I}$, with $\sigma$ increasing.  We again see there is a large range for the noise over which the local solver can become trapped in a local minimum while the global solver remains tight; in fact, even at very low noise levels it is quite easy to have the local solver become trapped.  With $K = 21$ poses in the trajectory, the local solver took on average \change{$0.1928s$ while the SDP solver took on average $13.42s$.}

There are also a few noteworthy differences in the continuous-time experiment as compared to the discrete-time case.  First, the log \ac{SVR} values are quite a bit lower in Figure~\ref{fig:cayley_continuum_noise} as compared to Figure~\ref{fig:cayley_trajest3_noise}.  This seems to be mainly due to the fact that we are now using a sparser set of measurements.  Our continuous-time experiments had pose measurements only at the start, middle, and end, whereas the discrete-time case had pose measurements at every timestep.  It is known that a sparser measurement graph can impact \ac{SDP} tightness \citep{holmes23}.  \change{Second, the low-noise test cases experienced some numerical issues with getting the \ac{SDP} solver to reliably converge.  It seems this is related to matrix conditioning resulting from the fact that we are marginalizing out the $\mbs{\varpi}_k$ variables before solving the \ac{SDP}.  We found it was necessary to adjust the scaling of the $\mbf{Q}$ matrix in~\eqref{eq:sdp2} to get reliable solutions.  Despite our best efforts, we see that there was one out of $100$ test cases at the $\sigma = 0.01$ noise level where the \ac{SDP} failed to solve.  On the log \ac{SVR} plot this shows as a red plus sign at $0$ and results in the success rate of the \ac{SDP} solver being $0.99$ instead of $1$.  Still, we {\em know} that the solver failed and therefore not to trust the answer. } Overall, we still have log \ac{SVR} values \change{almost always } above $5$ up to about $\sigma = 1$, which makes the solution practical\change{; it } is easy for the local solver to become stuck in local minima in this noise range.

\section{Conclusion and Future Work}
\label{sec:conclusion}

We have presented several new convex relaxations for pose and rotation estimation problems based on the Cayley map.  Our results indicate that for small problem sizes, we can successfully achieve global optimality with realistic amounts of noise and even with measurement sparsity in the case of continuous-time trajectory estimation.  \change{In each of the experiments, we indicated that covariance of the error associated with each pose measurement cost term is $\sigma^2 \mathbf{I}$.  In other words, $\sigma$ is the standard deviation of the measurement noise.  In the case of rotational degrees of freedom, $\sigma = 0.5$ is already shown in Figure~\ref{fig:maps} to represent quite a lot of rotational uncertainty, indeed more than typically occurs in practice.  Since the standard deviation is in radians, this implies that uncertainty spreads out over a large part of a full circle with $\sigma = 0.5$.  The fact that our convex relaxations empirically remain tight (\change{ensuring } global optimality) beyond the $\sigma = 0.5$ level (often beyond $\sigma = 1$) means our technique works over most practical situations.  For translational degrees of freedom, $\sigma$ will have units of distance.  The trajectories in the examples have poses that are spaced one distance unit apart so when the noise on the measurement of one of these poses is $\sigma = 0.5$ distance units, that is again quite a lot of noise in comparison to the spacing of the poses.  The implication is again that our convex relaxations remain tight (\change{ensuring } global optimality) for most practical situations.

While our results are promising, }we are still relying on off-the-shelf solvers once our problem has been converted to an \ac{SDP}, which means that we will not be able to scale up to extremely large state sizes.
To scale up, there are a few possibilities that we could explore.  First, perhaps we might be satisfied with merely certifying our local solver solutions.  Other works have focussed on this.  The challenge is that in most of the problems of this paper, we require redundant constraints to tighten our \acp{SDP}.  This means that we do not meet the technical condition of \ac{LICQ} \citep{boyd04}.  It turns out that this makes it more challenging to calculate an optimality certificate.  \citet{yang20} is a practical example where a certificate has been constructed for this type of situation, but there are still scaling issues.  Another possibility is to solve our problems globally using the approach of \citet{burer05} (studied more recently by \citet{boumal16}).  This was exploited with very impressive results by \citet{rosen19, dellaert20}; however, these problems enjoyed \ac{LICQ}.  \change{To our knowledge, this approach has not been applied to problems in robotics with redundant constraints. }  Or, perhaps generic \ac{SDP} solvers can be made to better exploit problem-specific structure \change{(e.g., chordal sparsity). }  We plan to explore these and other ways of scaling global optimality for a larger range of state estimation problems.

\input{cayopt.bbl}

\appendix

\change{
\section{Comparing Exponential and Cayley Distributions}
\label{sec:distributions}

We use this section to analyze how similar our proposed Gaussian-like distribution via the Cayley map is to one via the exponential map.  

Consider two distributions for rotations, $\mbf{C}_1$ and $\mbf{C}_2$, defined according to
\beqn{}
\mbf{C}_1 & = & \exp\left(\mbs{\phi}_1^\wdg \right) \bar{\mbf{C}}, \quad \mbs{\phi}_1 \sim \mathcal{N} \left( \mbf{0}, \mbs{\Sigma}_1 \right), \\
\mbf{C}_2 & = & \cay\left(\mbs{\phi}_2^\wdg \right) \bar{\mbf{C}}, \quad \mbs{\phi}_2 \sim \mathcal{N} \left( \mbf{0}, \mbs{\Sigma}_2 \right).
\eeqn
To ensure that $\mbf{C}_1 = \mbf{C}_2$ (in every instance), \citet{bauchau03} show that we require
\beqn{axisangle}
\mbs{\phi}_1 & = & \phi_1 \, \mbf{a} \; = \; \varphi \, \mbf{a}, \\
\mbs{\phi}_2 & = & \phi_2 \, \mbf{a} \; = \;  2 \tan \left( \frac{\varphi}{2} \right) \, \mbf{a},
\eeqn
in terms of a common axis of rotation, $\mbf{a}$, and angle of rotation, $\varphi$.  Notationally, we define
 $\phi_1 = || \mbs{\phi}_1 || = \varphi$ and $\phi_2 = || \mbs{\phi}_2 || = 2 \tan \left( \frac{\varphi}{2} \right)$.
Rearranging, we therefore require 
\begin{equation}\label{eq:phi1phi2}
\mbs{\phi}_2 = \frac{\tan\left( \frac{\phi_1}{2}\right)}{\frac{\phi_1}{2}} \mbs{\phi}_1
\end{equation}
to relate $\mbs{\phi}_1$ and $\mbs{\phi}_2$ directly.

We can use~\eqref{eq:phi1phi2} to relate the distribution for $\mbs{\phi}_2$ to that of $\mbs{\phi}_1$, which will in turn make the distribution for $\mbf{C}_2$ approximately match that of $\mbf{C}_1$.  Specifically, we will try to match the moments of the left and right sides of~\eqref{eq:phi1phi2}.  Moment matching is known to minimize a version of Kullback-Liebler divergence between two distributions \citep{bishop06}. 

Expanding the $\tan$ function in~\eqref{eq:phi1phi2} using a Taylor series we can write
\begin{equation}
\label{eq:exp2cay}
\mbs{\phi}_2 = \left( 1 + \frac{1}{12} \mbs{\phi}_1^T \mbs{\phi}_1 + \frac{1}{120} (\mbs{\phi}_1^T \mbs{\phi}_1)^2 + \cdots \right) \mbs{\phi}_1.
\end{equation}
We have only odd powers of $\mbs{\phi}_1$ appearing on the right side of this expression so we immediately see that we want
\begin{equation}
E[ \mbs{\phi}_2 ] = \mbf{0}.
\end{equation}
This does not involve any approximation.  We only need to use that $\mbs{\phi}_1$ follows a Gaussian distribution, which has all its odd moments zero.

For the covariance matrix, $\mbs{\Sigma}_2 = E\left[ \mbs{\phi}_2 \mbs{\phi}_2^T \right]$, we can insert~\eqref{eq:exp2cay} and truncate after a number of terms to match the required covariance approximately.  For example, keeping terms out to quartic in $\mbs{\phi}_1$ we have
\begin{multline}
\mbs{\Sigma}_2 \approx E\left[\mbs{\phi}_1 \mbs{\phi}_1^T \right] + \frac{1}{6} E\left[ \mbs{\phi}_1 \mbs{\phi}_1^T \mbs{\phi}_1 \mbs{\phi}_1^T \right] \\ = \mbs{\Sigma}_1 + \frac{1}{6} \left( \mbox{tr}(\mbs{\Sigma}_1) \mbf{I} + 2\mbs{\Sigma}_1 \right) \mbs{\Sigma}_1,
\end{multline}
where we have employed Isserlis' theorem \citep[\S 2.2.17]{barfoot_ser24} to compute the expectation on the right.  Thus defining
\begin{equation}
\label{eq:secondorder}
\mbs{\phi}_2 \sim \mathcal{N}\biggl( \mbf{0}, \underbrace{\mbs{\Sigma}_1 + \frac{1}{6} \left( \mbox{tr}(\mbs{\Sigma}_1)\mbf{I} + 2\mbs{\Sigma}_1 \right) \mbs{\Sigma}_1}_{\mbs{\Sigma}_2} \biggr)
\end{equation}
will result in $\mbf{C}_1$ and $\mbf{C}_2$ approximately having the same distribution.  We will refer to the approximation $\mbs{\Sigma}_2 \approx \mbs{\Sigma}_1$ as `first order' (in $\mbs{\Sigma}_1$) and the better version in~\eqref{eq:secondorder} as `second order' (in $\mbs{\Sigma}_1$).  More terms in the approximation could be computed depending on the desired level of accuracy.

\begin{figure}[t]
\centering
\includegraphics[width=0.48\textwidth]{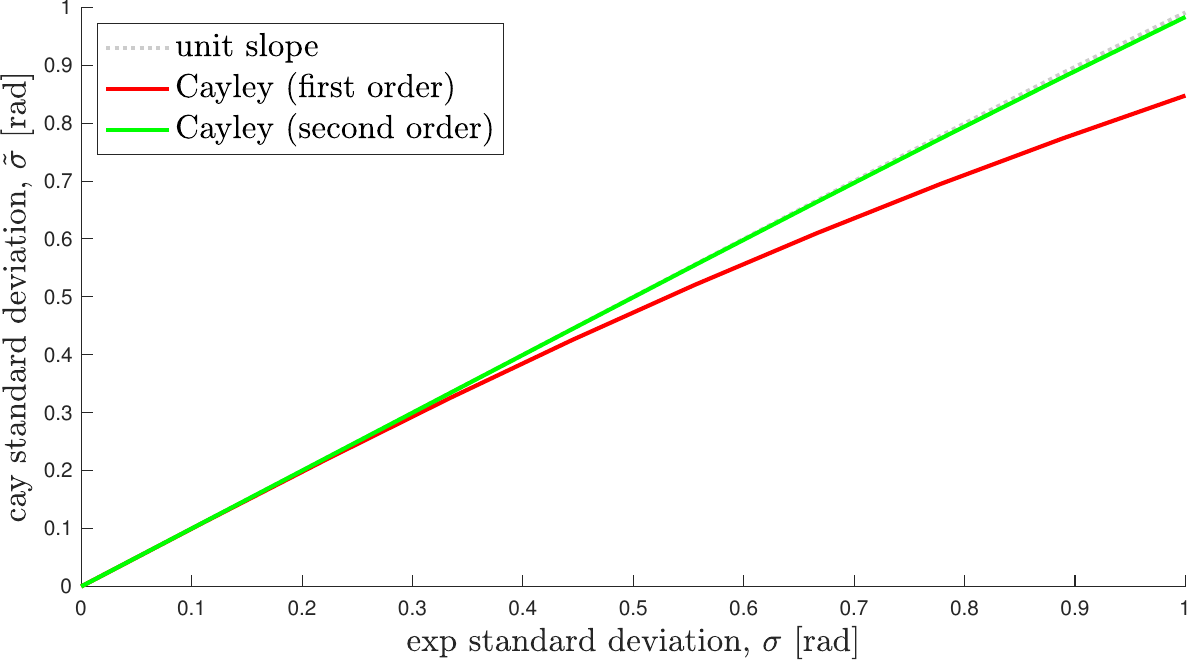}
\caption{Comparison of the resulting standard deviation of rotational uncertainty between the $\exp$ map (horizontal axis) and $\cay$ map (vertical) axis.  Red (first order) shows the naive approach of letting $\mbs{\Sigma}_2 = \mbs{\Sigma}_1$ and green (second order) shows the improvement offered by~\eqref{eq:secondorder}.}
\label{fig:maps_std}
\end{figure}

To visualize our approach, we can provide a one-dimensional example.  Consider $\mbs{\Sigma}_1 = \mbox{diag}(\sigma_1^2,0,0)$, so the axis of rotation, $\mbf{a} = \bbm 1 & 0 & 0 \ebm^T$, is held fixed.  Then the distributions for $\phi_1$ and $\phi_2$ will be
\begin{equation}\label{eq:phidist}
p_1(\phi_1) = \mathcal{N}(0,\sigma_1^2), \quad p_2(\phi_2) = \mathcal{N}(0, \sigma_2^2),
\end{equation}
where either $\sigma_2^2 = \sigma_1^2$ (first order) or $\sigma_2^2 = \sigma_1^2 + \frac{1}{2} \sigma_1^4$ (second order using~\eqref{eq:secondorder}).
If we map $\phi_1$ and $\phi_2$ through $\exp$ and $\cay$, respectively, we can produce densities on the actual rotational angle, $\varphi$: 
\beqn{}
\exp & : & \scalebox{0.9}{$p_1(\varphi) = \frac{1}{\sqrt{2 \pi \sigma_1^2}} \exp\left( -\frac{1}{2} \frac{\varphi^2}{\sigma_1^2} \right)$}, \\
\cay & : & \scalebox{0.9}{$p_2(\varphi) = \frac{1}{\sqrt{2 \pi \sigma_2^2}}\exp\left( -\frac{1}{2} \frac{(2 \tan(\frac{\varphi}{2}))^2}{\sigma_2^2} \right) \sec^2\left(\frac{\varphi}{2}\right)$}. \qquad
\eeqn
We arrive at these by inserting the substitutions from~\eqref{eq:axisangle} into the distributions in~\eqref{eq:phidist}.  The $\exp$ distribution is straightforward since $\phi_1 = \varphi$, which results in a Gaussian.  The $\cay$ distribution is slightly less obvious so we show the steps.  We first note that
\begin{equation}
\phi_2 = 2 \tan\left(\frac{\varphi}{2}\right) \quad \Rightarrow  \quad d\phi_2 = \sec^2\left(\frac{\varphi}{2}\right) \, d\varphi.
\end{equation} 
Then, to continue to satisfy the axiom of total probability we require
\begin{multline}
1 = \int \underbrace{\frac{1}{\sqrt{2 \pi \sigma_2^2}} \exp\left( -\frac{1}{2} \frac{\phi_2^2}{\sigma_2^2} \right)}_{p_2(\phi_2)} \, d\phi_2 \\ = \int \underbrace{\frac{1}{\sqrt{2 \pi \sigma_2^2}} \exp\left( -\frac{1}{2} \frac{(2 \tan(\frac{\varphi}{2}))^2}{\sigma_2^2} \right) \sec^2\left(\frac{\varphi}{2}\right)}_{p_2(\varphi)} \, d\varphi.
\end{multline}
We see that $p_2(\varphi)$ is no longer Gaussian in this case.  However, the idea is that the variances of $p_1(\varphi)$ and $p_2(\varphi)$ should be close when we compute $\sigma_2$ using~\eqref{eq:secondorder}.  Figure~\ref{fig:maps} provides examples when $\sigma_1 = 0.2$ [rad] and $\sigma_1 = 0.5$ [rad]; despite the latter being a fairly large rotational uncertainty, we see the distributions match quite well; their variances are almost identical.

Figure~\ref{fig:maps_std} shows the standard deviation, $\tilde{\sigma}$, of $\varphi$ using the Cayley map distribution (numerically computed from $p_2(\varphi)$) as $\sigma = \sigma_1$ is increased.  We see the standard deviations, $\sigma$ and $\tilde{\sigma}$, match quite well up to a large amount of rotational uncertainty.  These results suggest that replacing instances of the exponential map with the Cayley map may have little effect; depending on the amount of rotational uncertainty, we may choose to `correct' the covariance using~\eqref{eq:secondorder} or we could even include more terms to improve the approximation.
}

\section{Local Solvers}
\label{sec:localsolvers}

As the local solvers are not the main subject of the paper, we use this appendix to explain how these were implemented for each of the estimation problems.  We begin by outlining the required Lie-group Jacobians.

\subsection{Jacobians for Rotations and Poses}

In order to devise local optimization algorithms that work with Lie groups, we require the (inverse) Jacobians for $SO(3)$ and $SE(3)$.  For example, for rotations parameterized through the exponential map, $\mbf{C}(\mbs{\phi}) = \exp(\mbs{\phi}^\wdg)$, the inverse left Jacobian expression is well known to be \citep{barfoot_ser24}
\begin{equation}\label{eq:expSO3jacinv}
\mbf{J}_e(\mbs{\phi})^{-1} = \frac{\phi}{2}\cot\frac{\phi}{2} \mbf{I}  - \frac{\phi}{2} \mbf{a}^\wdg + \left( 1 - \frac{\phi}{2}\cot\frac{\phi}{2} \right) \mbf{a}\mbf{a}^T.
\end{equation}
The rotational kinematics can thus be written in one of two ways:
\begin{equation}
\dot{\mbf{C}}(\mbs{\phi}) = \mbs{\om}^\wdg \mbf{C}(\mbs{\phi}) \quad \Leftrightarrow \quad \dot{\mbs{\phi}} = \mbf{J}_e(\mbs{\phi})^{-1} \mbs{\om},
\end{equation}
where $\mbs{\om}$ is the angular velocity.  Naturally, the expression on the right is subject to singularities.  Considering an infinitesimal increment of time, the same Jacobian thus allows us to approximate the compounding of two exponentiated vectors $\mbs{\phi}_1, \mbs{\phi}_2 \in \mathbb{R}^3$ as
\begin{equation}
\exp( \mbs{\phi}_1^\wdg ) \, \exp( \mbs{\phi}_2^\wdg ) \approx \exp\left( \left( \mbf{J}_e(\mbs{\phi}_2)^{-1} \mbs{\phi}_1 + \mbs{\phi}_2\right)^\wdg \right),
\end{equation} 
where $\mbs{\phi}_1$ is assumed to be `small'.  A similar situation exists for $SE(3)$ poses parametrized by the exponential map, $\mbf{T}(\mbs{\xi}) = \exp(\mbs{\xi}^\wdg)$.  The kinematics can be written in one of two ways:
\begin{equation}
\dot{\mbf{T}}(\mbs{\xi}) = \mbs{\varpi}^\wdg \mbf{T}(\mbs{\xi}) \quad \Leftrightarrow \quad \dot{\mbs{\xi}} = \mbs{\mathcal{J}}_e(\mbs{\xi})^{-1} \mbs{\varpi},
\end{equation}
where $\mbs{\varpi}$ is the generalized velocity or `twist' and $\mbs{\mathcal{J}}_e(\mbs{\xi})^{-1}$ is the Jacobian (expression can be found in \citet{barfoot_ser24}).  Again, this allows us to approximate the compounding of two exponentiated vectors $\mbs{\xi}_1, \mbs{\xi}_2 \in \mathbb{R}^6$ as
\begin{equation}
\exp( \mbs{\xi}_1^\wdg ) \, \exp( \mbs{\xi}_2^\wdg ) \approx \exp\left( \left( \mbs{\mathcal{J}}_e(\mbs{\xi}_2)^{-1} \mbs{\xi}_1 + \mbs{\xi}_2\right)^\wdg \right),
\end{equation} 
where $\mbs{\xi}_1$ is assumed to be `small'.

To work with the Cayley map (instead of the exponential map), we will require similar Jacobian expressions, which are summarized neatly by \citet{muller21}.  In our notation, for rotations parameterized by the Cayley map, $\mbf{C}(\mbs{\phi}) = \cay(\mbs{\phi}^\wdg)$, the inverse left Jacobian expression is
\begin{equation}
\mbf{J}_c(\mbs{\phi})^{-1} = \mbf{I} - \frac{1}{2} \mbs{\phi}^\wdg + \frac{1}{4} \mbs{\phi} \mbs{\phi}^T,
\end{equation}
which is appealing in that it does not involve any trigonometric functions compared to~\eqref{eq:expSO3jacinv}.  In passing, the left Jacobian itself is the tidy expression,
\begin{equation}
\mbf{J}_c(\mbs{\phi}) = \frac{1}{1 +\frac{1}{4} \mbs{\phi}^T\mbs{\phi}} \left( \mbf{I} + \frac{1}{2} \mbs{\phi}^\wdg \right).
\end{equation}
The rotational kinematics can be written as
\begin{equation}
\dot{\mbf{C}}(\mbs{\phi}) = \mbs{\om}^\wdg \mbf{C}(\mbs{\phi}) \quad \Leftrightarrow \quad \dot{\mbs{\phi}} = \mbf{J}_c(\mbs{\phi})^{-1} \mbs{\om},
\end{equation}
and the compounding approximation for vectors $\mbs{\phi}_1, \mbs{\phi}_2 \in \mathbb{R}^3$ is
\begin{equation}\label{eq:cayrotcomp}
\cay( \mbs{\phi}_1^\wdg ) \, \cay( \mbs{\phi}_2^\wdg ) \approx \cay\left( \left( \mbf{J}_c(\mbs{\phi}_2)^{-1} \mbs{\phi}_1 + \mbs{\phi}_2\right)^\wdg \right),
\end{equation} 
where $\mbs{\phi}_1$ is assumed to be `small'.  Finally, for poses parameterized by the Cayley map, $\mbf{T}(\mbs{\xi}) = \cay(\mbs{\xi}^\wdg)$, the kinematics are
\begin{equation}
\dot{\mbf{T}}(\mbs{\xi}) = \mbs{\varpi}^\wdg \mbf{T}(\mbs{\xi}) \quad \Leftrightarrow \quad \dot{\mbs{\xi}} = \mbs{\mathcal{J}}_c(\mbs{\xi})^{-1} \mbs{\varpi},
\end{equation}
where $\mbs{\varpi}$ is the generalized velocity or `twist' and $\mbs{\mathcal{J}}_c(\mbs{\xi})^{-1}$ is the inverse Jacobian:
\begin{subequations}
\begin{gather}
\mbs{\mathcal{J}}_c(\mbs{\xi})^{-1} = \mbf{I} - \frac{1}{2} \mbs{\xi}^\Wdg + \frac{1}{4} \mbs{\Lambda}(\mbs{\xi}\mbs{\xi}^T), \\ \mbs{\Lambda}(\mbs{\xi}\mbs{\xi}^T) = \bbm \mbf{0} & \mbs{\phi}^\wdg \mbs{\rho}^\wdg \\ \mbf{0} & \mbs{\phi} \mbs{\phi}^T \ebm.
\end{gather}
\end{subequations}
Again, we have a nice expression devoid of any trigonometric functions.  The compounding approximation for two vectors $\mbs{\xi}_1, \mbs{\xi}_2 \in \mathbb{R}^6$ is then
\begin{equation}
\cay( \mbs{\xi}_1^\wdg ) \, \cay( \mbs{\xi}_2^\wdg ) \approx \cay\left( \left( \mbs{\mathcal{J}}_c(\mbs{\xi}_2)^{-1} \mbs{\xi}_1 + \mbs{\xi}_2\right)^\wdg \right),
\end{equation} 
where $\mbs{\xi}_1$ is assumed to be `small'.

\subsection{Rotation Averaging}

To devise a local solver for~\eqref{eq:rotavprob}, we begin with an initial guess, $\mbf{C}_{\rm op}$, and perturb it according to
\begin{equation}\label{eq:rotavpert}
\mbf{C} = \cay(\mbs{\psi}^\wdg) \mbf{C}_{\rm op},
\end{equation}
where $\mbs{\psi} \in \mathbb{R}^3$ is a small unknown perturbation.  This ensures that $\mbf{C} \in SO(3)$ during the optimization process.
The error in the cost function is then
\begin{multline}
\mbs{\phi}_m = \cay^{-1}\left(\mbf{C}\tilde{\mbf{C}}_m^T \right)^\vee = \cay^{-1}\left(\cay(\mbs{\psi}^\wdg) \mbf{C}_{\rm op}\tilde{\mbf{C}}_m^T \right)^\vee \\ \approx \mbf{J}_c( \mbs{\phi}_{{\rm op},m})^{-1} \mbs{\psi} + \mbs{\phi}_{{\rm op},m},
\end{multline}
where we have used~\eqref{eq:cayrotcomp} to arrive at the approximation and defined $\mbs{\phi}_{{\rm op},m} =  \cay^{-1}\left(\mbf{C}_{\rm op}\tilde{\mbf{C}}_m^T \right)^\vee$.  A quadratic approximation to the cost function in~\eqref{eq:rotavprob} (in terms of the perturbation) is thus
\begin{multline}
\sum_{m=1}^M \left( \mbf{J}_c( \mbs{\phi}_{{\rm op},m})^{-1} \mbs{\psi} + \mbs{\phi}_{{\rm op},m} \right)^T \\ \times \; \mbf{W}_m \left( \mbf{J}_c( \mbs{\phi}_{{\rm op},m})^{-1} \mbs{\psi} + \mbs{\phi}_{{\rm op},m} \right).
\end{multline}
To minimize this with respect to $\mbs{\psi}$, we take the derivative and set to zero for an optimum, resulting in the linear system of equations
\begin{multline}
\left( \sum_{m=1}^M \mbf{J}_c( \mbs{\phi}_{{\rm op},m})^{-T} \mbf{W}_m \mbf{J}_c( \mbs{\phi}_{{\rm op},m})^{-1} \right) \mbs{\psi}^\star \\ = \sum_{m=1}^M \mbf{J}_c( \mbs{\phi}_{{\rm op},m})^{-T} \mbf{W}_m \mbs{\phi}_{{\rm op},m},
\end{multline}
which we solve for the optimal perturbation, $\mbs{\psi}^\star$.  We then apply this optimal perturbation to the initial guess according to~\eqref{eq:rotavpert}, and iterate the process to convergence.  This is effectively Gauss-Newton adapted to work for $SO(3)$ and will converge locally in practice.

\subsection{Pose Averaging}

A very similar local solver can be devised for~\eqref{eq:poseavprob}.  We begin with an initial guess, $\mbf{T}_{\rm op}$, and perturb it according to
\begin{equation}\label{eq:poseavpert}
\mbf{T} = \cay(\mbs{\ep}^\wdg) \mbf{T}_{\rm op},
\end{equation}
where $\mbs{\ep} \in \mathbb{R}^6$ is a small unknown perturbation.  This ensures that $\mbf{T} \in SE(3)$ during the optimization process.  A quadratic approximation to the cost function in~\eqref{eq:poseavprob} (in terms of the perturbation) is thus
\begin{multline}
\sum_{m=1}^M \left( \Jbig_c( \mbs{\xi}_{{\rm op},m})^{-1} \mbs{\ep} + \mbs{\xi}_{{\rm op},m} \right)^T \\ \times \; \mbf{W}_m \left( \Jbig_c( \mbs{\xi}_{{\rm op},m})^{-1} \mbs{\ep} + \mbs{\xi}_{{\rm op},m} \right),
\end{multline}
where $\mbs{\xi}_{{\rm op},m} = \cay^{-1}\left(\mbf{T}_{\rm op}\tilde{\mbf{T}}_m^{-1} \right)^\vee$.
To minimize this with respect to $\mbs{\ep}$, we take the derivative and set to zero for an optimum, resulting in the linear system of equations
\begin{multline}
\left( \sum_{m=1}^M \Jbig_c( \mbs{\xi}_{{\rm op},m})^{-T} \mbf{W}_m \Jbig_c( \mbs{\xi}_{{\rm op},m})^{-1} \right) \mbs{\ep}^\star \\ = \sum_{m=1}^M \Jbig_c( \mbs{\xi}_{{\rm op},m})^{-T} \mbf{W}_m \mbs{\xi}_{{\rm op},m},
\end{multline}
which we solve for the optimal perturbation, $\mbs{\ep}^\star$.  We then apply this optimal perturbation to the initial guess according to~\eqref{eq:poseavpert}, and iterate the process to convergence.  This is effectively Gauss-Newton adapted to work for $SE(3)$ and will converge locally in practice.

\subsection{Discrete-Time Trajectory Estimation}

To devise a local solver for~\eqref{eq:trajestprob}, we begin with initial guesses, $\mbf{T}_{\rm op,k}$, and perturb them according to
\begin{equation}\label{eq:trajestpert}
\mbf{T}_k = \cay(\mbs{\ep}_k^\wdg) \mbf{T}_{\rm op,k},
\end{equation}
where $\mbs{\ep}_k \in \mathbb{R}^6$ are small unknown perturbations.  This ensures that $\mbf{T}_k \in SE(3)$ during the optimization process.  A quadratic approximation to the cost function in~\eqref{eq:trajestprob} (in terms of the perturbations) is thus
\begin{multline}
\scalebox{0.85}{
$\sum_{k=1}^K \left( \Jbig_c( \mbs{\xi}_{{\rm op},k})^{-1} \mbs{\ep}_k + \mbs{\xi}_{{\rm op},k} \right)^T \mbf{W}_k \left( \Jbig_c( \mbs{\xi}_{{\rm op},k})^{-1} \mbs{\ep}_k + \mbs{\xi}_{{\rm op},k} \right)$} \\ 
\scalebox{0.85}{$+ \; \sum_{k=1}^{K-1} \left( \Jbig_c( \mbs{\xi}_{{\rm op},k+1,k})^{-1} \left( \mbs{\ep}_{k+1} - \Tbig_{{\rm op},k+1,k} \mbs{\ep}_k \right) + \mbs{\xi}_{{\rm op},k+1,k} \right)^T$} \\ 
\scalebox{0.85}{$\times \; \mbf{W}_{k+1,k} \left( \Jbig_c( \mbs{\xi}_{{\rm op},k+1,k})^{-1} \left( \mbs{\ep}_{k+1} - \Tbig_{{\rm op},k+1,k} \mbs{\ep}_k \right) + \mbs{\xi}_{{\rm op},k+1,k} \right)$},  \\
\end{multline}
where $\mbs{\xi}_{{\rm op},k} = \cay^{-1}\left(\mbf{T}_{{\rm op},k}\tilde{\mbf{T}}_k^{-1} \right)^\vee$,  $\mbs{\xi}_{{\rm op},k+1,k} = \cay^{-1}\left(\mbf{T}_{{\rm op},k+1}\mbf{T}_{{\rm op},k}^{-1}\tilde{\mbf{T}}_{k+1,k}^{-1} \right)^\vee$, and $\Tbig_{{\rm op},k+1,k} = \mbox{Ad}\left( \mbf{T}_{{\rm op},k+1} \mbf{T}_{{\rm op},k}^{-1} \right)$.  We can tidy this up by defining
\begin{subequations}
\begin{gather}
\scalebox{0.72}{
$\mbf{H}  = \bbm -\Jbig_c( \mbs{\xi}_{{\rm op},1})^{-1} & \mbf{0} & \cdots & \mbf{0} \\  \Jbig_c( \mbs{\xi}_{{\rm op},2,1})^{-1} \Tbig_{{\rm op},2,1} & - \Jbig_c( \mbs{\xi}_{{\rm op},2,1})^{-1} & \cdots & \mbf{0} \\ \mbf{0} & -\Jbig_c( \mbs{\xi}_{{\rm op},2})^{-1} & \cdots & \mbf{0} \\ \mbf{0} &  \Jbig_c( \mbs{\xi}_{{\rm op},3,2})^{-1} \Tbig_{{\rm op},3,2} & \cdots & \mbf{0} \\ \vdots & \vdots & \ddots & \vdots \\ \mbf{0} & \mbf{0} & \cdots & -\Jbig_c( \mbs{\xi}_{{\rm op},K})^{-1} \ebm$}, \\ \mbs{\xi}_{\rm op} = \bbm \mbs{\xi}_{{\rm op},1} \\ \mbs{\xi}_{{\rm op},2,1} \\ \mbs{\xi}_{{\rm op},2} \\ \mbs{\xi}_{{\rm op},3,2} \\ \vdots \\ \mbs{\xi}_{{\rm op},K} \ebm, \quad \mbs{\ep} = \bbm \mbs{\ep}_1 \\ \mbs{\ep}_2 \\ \vdots \\ \mbs{\ep}_K \ebm, \\
\mbf{W} = \mbox{diag}\left( \mbf{W}_1, \mbf{W}_{2,1}, \mbf{W}_2, \ldots, \mbf{W}_{K,K-1}, \mbf{W}_K \right),
\end{gather}
\end{subequations}
whereupon the quadratic approximation to the cost function can be written as
\begin{equation}
\left( \mbs{\xi}_{\rm op} - \mbf{H} \mbs{\ep}\right)^T \mbf{W} \left( \mbs{\xi}_{\rm op} - \mbf{H} \mbs{\ep}\right).
\end{equation}
Minimizing this with respect to $\mbs{\ep}$ results in a linear system of equations:
\begin{equation}
\mbf{H}^T \mbf{W} \mbf{H} \, \mbs{\ep}^\star = \mbf{H}^T \mbf{W} \mbs{\xi}_{\rm op}.
\end{equation}
We solve for the optimal perturbations, $\mbs{\ep}^\star$, apply them to the initial guess according to~\eqref{eq:trajestpert}, and iterate the process to convergence.  This is again Gauss-Newton adapted to work for $SE(3)$ and will converge locally in practice.  \change{We found a line search was needed for this problem to ensure smooth convergence \citep{nocedal1999numerical}.}

\subsection{Continuous-Time Trajectory Estimation}

To devise a local solver for~\eqref{eq:cttrajestprob}, we begin with initial guesses, $\left\{ \mbf{T}_{\rm op,k}, \mbs{\varpi}_{{\rm op},k} \right\}$, and perturb them according to
\begin{equation}\label{eq:cttrajestpert}
\mbf{T}_k = \cay(\mbs{\ep}_k^\wdg) \mbf{T}_{\rm op,k}, \quad \mbs{\varpi} = \mbs{\varpi}_{{\rm op},k} + \mbs{\eta}_k,
\end{equation}
where $\mbs{\ep}_k, \mbs{\eta}_k \in \mathbb{R}^6$ are small unknown perturbations.  This ensures that $\mbf{T}_k \in SE(3)$ during the optimization process.  A quadratic approximation to the cost function in~\eqref{eq:cttrajestprob} (in terms of the perturbations) is thus
\begin{multline}
\sum_{k=1}^K \left(  \mbs{\xi}_{{\rm op},k} - \mbf{G}_k \mbf{x}_k \right)^T \mbf{W}_k \left(  \mbs{\xi}_{{\rm op},k} - \mbf{G}_k \mbf{x}_k \right) \\ + (\pri{\mbs{\varpi}}_1 - \mbs{\varpi}_{{\rm op},1} - \mbf{E}_1 \mbf{x}_1)^T \mbf{Q}^{-1}_1 (\pri{\mbs{\varpi}}_1 - \mbs{\varpi}_{{\rm op},1} - \mbf{E}_1 \mbf{x}_1)\\ + \sum_{k=1}^{K-1} \left( \mbf{e}_{{\rm op},k+1,k} + \mbf{F}_{k} \mbf{x}_{k}  - \mbf{E}_{k+1} \mbf{x}_{k+1} \right)^T \qquad  \\ \qquad \times \; \mbf{Q}_{k+1,k}^{-1} \left( \mbf{e}_{{\rm op},k+1,k} + \mbf{F}_{k} \mbf{x}_{k}  - \mbf{E}_{k+1} \mbf{x}_{k+1} \right) ,
\end{multline}
where 
\begin{subequations}
\begin{gather}
\mbf{x}_k = \bbm \mbs{\ep}_k \\ \mbs{\eta}_k \ebm, \\ \mbf{e}_{{\rm op}, k+1,k} = \bbm (t_{k+1} - t_k) \mbs{\varpi}_{{\rm op},k} - \mbs{\xi}_{{\rm op},k+1,k} \\ \mbs{\varpi}_{{\rm op},k} - \mbs{\varpi}_{{\rm op}, k+1}  \ebm, \\ \mbf{G} = \bbm  -\Jbig_c( \mbs{\xi}_{{\rm op},k})^{-1} & \mbf{0} \ebm, \quad \mbf{E}_1 = \bbm \mbf{0} & \mbf{I} \ebm,  \\ \mbf{F}_{k} = \bbm \Jbig_c( \mbs{\xi}_{{\rm op},k+1,k})^{-1}\Tbig_{{\rm op},k+1,k} & (t_{k+1}-t_k) \mbf{I} \\ \mbf{0} & \mbf{I} \ebm, \\ \mbf{E}_{k+1} = \bbm  \Jbig_c( \mbs{\xi}_{{\rm op},k+1,k})^{-1} & \mbf{0} \\  \mbf{0} & \mbf{I}  \ebm
\end{gather}
\end{subequations}
with $\mbs{\xi}_{{\rm op},k} = \cay^{-1}\left(\mbf{T}_{{\rm op},k}\tilde{\mbf{T}}_k^{-1} \right)^\vee$,  $\mbs{\xi}_{{\rm op},k+1,k} = \cay^{-1}\left(\mbf{T}_{{\rm op},k+1}\mbf{T}_{{\rm op},k}^{-1} \right)^\vee$, and $\Tbig_{{\rm op},k+1,k} = \mbox{Ad}\left( \mbf{T}_{{\rm op},k+1} \mbf{T}_{{\rm op},k}^{-1} \right)$.  We can tidy this up by defining the following\footnote{The notation was chosen to match the continuous-time estimation chapter of \citet{barfoot_ser24}.}
\begin{subequations}
\begin{gather}
\mbf{F}^{-1} = \bbm 
\mbf{E}_1 & \mbf{0}  & \mbf{0} & \cdots & \mbf{0} & \mbf{0} \\
-\mbf{F}_1 & \mbf{E}_2 & \mbf{0} & \cdots & \mbf{0} & \mbf{0} \\
\mbf{0} & -\mbf{F}_2 & \mbf{E}_3  & \cdots & \mbf{0} & \mbf{0} \\
\vdots & \vdots & \vdots & \ddots & \vdots & \vdots \\
\mbf{0} & \mbf{0} & \mbf{0} & \cdots & -\mbf{F}_{K-1} & \mbf{E}_K
\ebm, \\ \mbf{Q} = \mbox{diag}\left( \mbf{Q}_1, \mbf{Q}_{2,1}, \ldots, \mbf{Q}_{K,K-1} \right),  \\ \quad 
\mbf{G} = \mbox{diag}\left( \mbf{G}_1, \mbf{G}_2, \ldots, \mbf{G}_K \right), \\ \mbf{R} = \mbox{diag}\left( \mbf{W}^{-1}_1, \mbf{W}^{-1}_2, \ldots, \mbf{W}^{-1}_K \right),  \\ \mbf{e}_{{\rm op},v} = \bbm \pri{\mbs{\varpi}}_1 - \mbs{\varpi}_{{\rm op},1} \\ \mbf{e}_{{\rm op},2,1} \\ \vdots \\ \mbf{e}_{{\rm op},K,K-1} \ebm, \quad \mbf{e}_{{\rm op},y} = \bbm \mbs{\xi}_{{\rm op},1} \\ \mbs{\xi}_{{\rm op},2} \\ \vdots \\ \mbs{\xi}_{{\rm op},K} \ebm, \\ \mbf{x} = \bbm \mbf{x}_1 \\ \mbf{x}_2 \\ \vdots \\ \mbf{x}_K \ebm,
\end{gather}
\end{subequations}
whereupon the cost function in~\eqref{eq:cttrajestprob} can be written as
\begin{multline}
\left(\mbf{e}_{{\rm op},y} - \mbf{G} \mbf{x} \right)^T \mbf{R}^{-1} \left( \mbf{e}_{{\rm op},y}- \mbf{G} \mbf{x} \right) \\ + \; \left( \mbf{e}_{{\rm op},v} - \mbf{F}^{-1} \mbf{x} \right)^T \mbf{Q}^{-1} \left( \mbf{e}_{{\rm op},v}- \mbf{F}^{-1} \mbf{x} \right).
\end{multline}
Minimizing this with respect to $\mbf{x}$ results in a linear system of equations:
\begin{multline}
\left( \mbf{G}^T \mbf{R}^{-1} \mbf{G} + \mbf{F}^{-T} \mbf{Q}^{-1} \mbf{F}^{-1} \right) \, \mbf{x}^\star \\ = \; \mbf{G}^T \mbf{R}^{-1} \mbf{e}_{{\rm op},y} + \mbf{F}^{-T} \mbf{Q}^{-1} \mbf{e}_{{\rm op},v}.
\end{multline}
We solve for the optimal perturbations, $\mbf{x}^\star$, unstack into $\mbs{\ep}_k^\star$ and $\mbs{\eta}_k^\star$, then apply them to the initial guess according to~\eqref{eq:cttrajestpert}, and iterate the process to convergence.  This is again Gauss-Newton adapted to work for $SE(3)$ and will converge locally in practice.   \change{We found a line search was needed for this problem to ensure smooth convergence \citep{nocedal1999numerical}. }
Note, if we do not have pose measurements at some of the estimation times, we can simply delete the appropriate rows/columns of $\mbf{G}$, $\mbf{R}$, and $\mbf{e}_{{\rm op},y}$. 

\end{document}

%% file: notation_header.tex
\newcommand{\bbm}{\begin{bmatrix}}
\newcommand{\ebm}{\end{bmatrix}}
\newcommand{\mbf}{\mathbf}
\newcommand{\mbs}[1]{{\boldsymbol{#1}}}
\def\ep{\epsilon}

\def\om{\omega}

\newcommand{\beq}{\begin{equation}}
\newcommand{\eeq}{\end{equation}}
\newcommand{\bdis}{\begin{displaymath}}
\newcommand{\edis}{\end{displaymath}}
\newcommand{\beqn}[1]{\begin{subequations}\label{eq:#1}\begin{eqnarray}}
\newcommand{\eeqn}{\end{eqnarray}\end{subequations}}

\newcommand{\pri}[1]{\check{#1}}
\newcommand{\wdg}{\wedge}

\newcommand{\Wdg}{\curlywedge}
\newcommand{\uWdg}{\curlyvee}
\newcommand{\Tbig}{\mbs{\mathcal{T}}}

\newcommand{\Jbig}{\mbs{\mathcal{J}}}

\newcommand{\cay}{\mbox{cay}}